\documentclass{article}

\PassOptionsToPackage{numbers, compress}{natbib}

\usepackage[preprint]{neurips_2025}

\usepackage[utf8]{inputenc} 
\usepackage[T1]{fontenc}    
\usepackage{hyperref}       
\usepackage{url}            
\usepackage{booktabs}       
\usepackage{amsfonts}       
\usepackage{nicefrac}       
\usepackage{microtype}      
\usepackage{xcolor}         
\usepackage{amsmath}
\usepackage{graphicx}
\usepackage{cleveref}
\usepackage{algorithm}
\usepackage{algorithmic}
\usepackage{multirow}
\usepackage{caption}
\usepackage[most]{tcolorbox}
\tcbuselibrary{listings, breakable, skins}
\usepackage{wrapfig}
\usepackage{subcaption}
\usepackage{booktabs}
\usepackage{longtable}
\usepackage{setspace}
\usepackage{caption}

\crefname{figure}{Figure}{Figures} 
\crefname{equation}{Equation}{Equations}
\crefname{table}{Table}{Tables}
\crefname{section}{Section}{Sections}

\title{Sentence-level Reward Model can Generalize Better for Aligning LLM from Human Preference}

\author{%
\textbf{Wenjie Qiu}$^{1,2}$\thanks{Equal contribution} \quad
\textbf{Yi-Chen Li}$^{1,2,3}$\footnotemark[1] \quad
\textbf{Xuqin Zhang}$^{1,2}$ \quad
\textbf{Tianyi Zhang}$^{1,2}$ \quad
\textbf{Yihang Zhang}$^{1,2}$ \\
\textbf{Zongzhang Zhang}$^{1,2}$\thanks{Corresponding author} \quad
\textbf{Yang Yu}$^{1,2,3}$ \\
$^1$National Key Laboratory for Novel Software Technology, Nanjing University, China \\
$^2$School of Artificial Intelligence, Nanjing University, China \\
$^3$Polixir Technologies, Nanjing, China
\vspace{-24pt}
}

\begin{document}

\maketitle

\begin{abstract}
  Learning reward models from human preference datasets and subsequently optimizing language models via reinforcement learning has emerged as a fundamental paradigm for aligning LLMs with human preferences. The performance of the reward model plays a crucial role in the effectiveness of alignment. Conventional reward models operate at a coarse-grained level, requiring the generation of a complete \emph{response} to obtain a reward value. Such sparse reward signals can introduce substantial difficulties for reinforcement learning. While recent efforts have attempted to learn \emph{token}-level reward models, the lack of explicit semantic information makes it difficult to model the credit of every individual token. In this paper, we introduce an an intermediate-grained reward model, proposing the assignment of rewards at the \emph{sentence} level. By segmenting the complete response into sentences and applying differential operations to reward outputs at the start and end positions of each sentence, we can effectively model the rewards of sentences. Moreover, a novel attention mechanism is introduced to aggregate the rewards of all sentences into a response-level reward, which allows it to be trained using the Bradley-Terry model. On established benchmarks, our proposed method yields a 2.7\% improvement over response-level reward models on RewardBench (for reward modeling evaluation) and surpasses all baselines on AlpacaEval (for overall alignment performance).
\end{abstract}

\section{Introduction}
\begin{figure*}[!htbp]
  \centering
    \includegraphics[width=0.98\linewidth,trim = 0.2cm 0.2cm 0.5cm 0.2cm, clip]{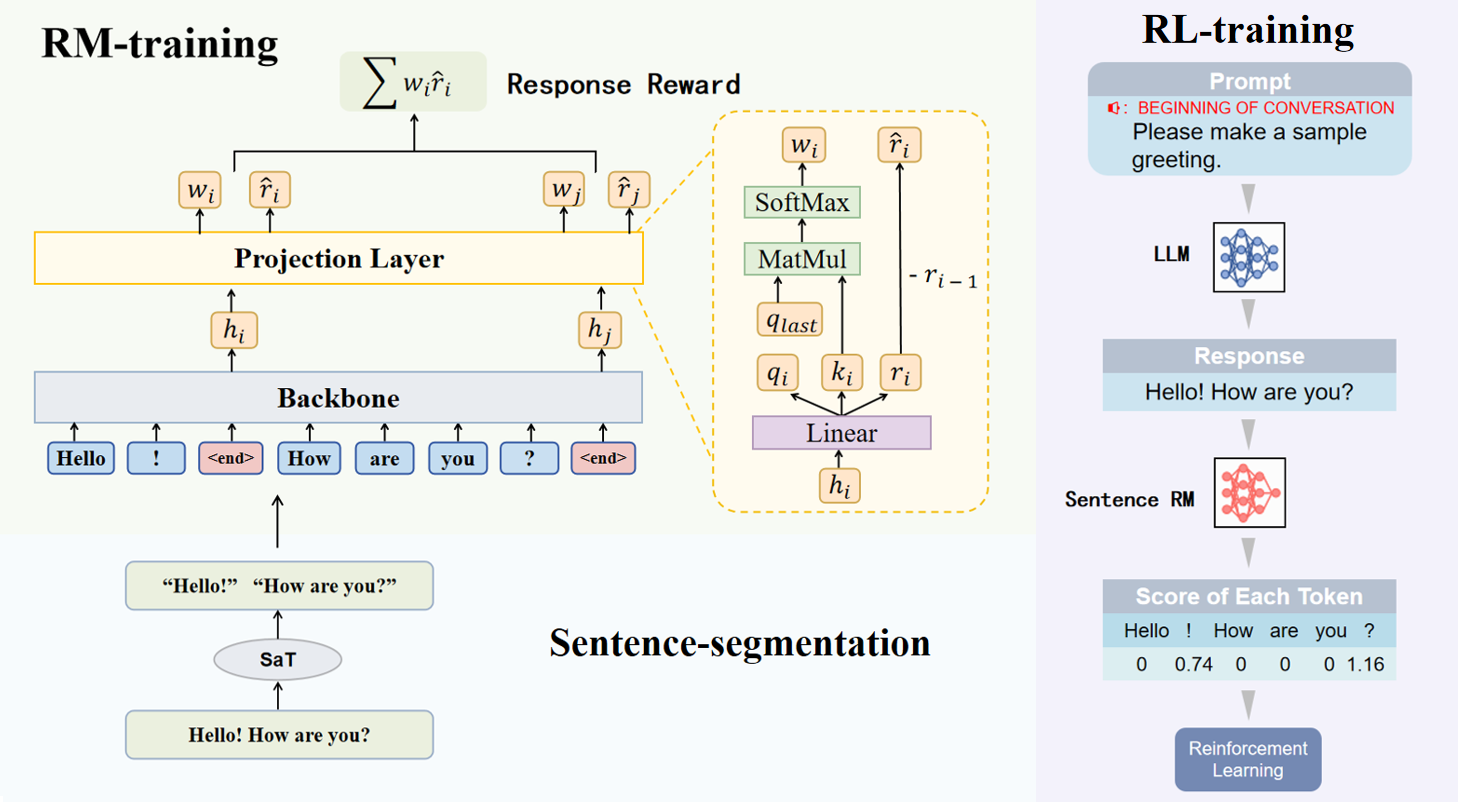}
  \caption{Framework of sentence-level reward modeling. Given an input response \((x, a_1, \cdots, a_T)\), a segmenter first identifies sentence boundaries. The projection layer then transforms the hidden states of these sentence boundary tokens into sentence-level rewards $\hat{r}_{i}$ and their corresponding weights $w_i$. The overall response-level reward is computed as a weighted sum of these sentence-level rewards. During RL training, the sentence-level reward model provides non-zero rewards at sentence boundaries, enabling the use of advanced reinforcement learning algorithms for LLM optimization.}
  \vspace{-18pt}
  \label{Framework}
\end{figure*}
\label{intro}
Reinforcement Learning from Human Feedback (RLHF) \citep[RLHF,][]{instruct-gpt} has emerged as a pivotal technique for enhancing the generative capabilities and aligning Large Language Models (LLMs) with human preferences. thereby contributing significantly to the success of prominent LLMs like ChatGPT~\citep{chatgpt} and LLaMA~\citep{llama}. The standard RLHF pipeline typically comprises two main stages: learning Reward Models (RMs) from human preference data, and fine-tuning the LLMs using Reinforcement Learning (RL) algorithms~\citep{instruct-gpt}. The learned RMs serve as a surrogate for human annotators, providing scalar rewards for responses generated by the LLMs. The reinforcement learning step employs algorithms such as REINFORCE~\citep{Reinforce} and PPO~\citep{PPO} to maximize the expected cumulative reward, thereby guiding LLMs towards generating outputs that better align with human preferences. The efficacy of the reward model profoundly impacts the overall performance of RLHF; a poorly performing reward model can severely degrade the LLM's output quality or even lead to catastrophic model collapse \citep{rm_scaling,rm_secrets}.

Significant research has focused on the development of effective reward models. Prior work has primarily employed response-level reward models, which assign a single reward score to an entire generated response \citep{BoN,instruct-gpt,llama}. While this approach simplifies reward model training, it introduces a granularity mismatch during the fine-tuning of LLMs. Since LLMs generate responses token by token, the response-level reward model only provides a reward signal at the final token (e.g., \texttt{EOS>}). This results in the challenge of sparse rewards ~\citep{HER,exploration_survey}. To mitigate this issue, recent efforts have explored the development of token-level reward models. However, from a linguistic perspective, assigning accurate reward scores to individual tokens is challenging as they often lack explicit semantic meaning in isolation. Even for human annotators, providing token-level preference labels is a highly challenging and often subjective task. From a training perspective, existing token-level reward modeling methods typically rely on either training a credit assignment model to propagate response-level preferences to individual tokens \citep{densefree,preference-grounded,RTO}, or leveraging AI systems for token-level preference annotation \citep{tlcr,beyondIL}. Both approaches present significant difficulties in training accurate and reliable token-level reward models.

Given these limitations, a natural question thus emerges: \textit{Can we identify a level of granularity that carries explicit semantic information while being finer-grained than an entire response?} Consider how humans evaluate an essay: we often break it down into smaller pieces, evaluate each piece, and then aggregate these evaluations into a final score. These "pieces" frequently correspond to individual sentences or groups of sentences in practice. Inspired by this observation, we propose reward modeling at the sentence level to introduce an intermediate granularity between token and response.

Our method's implementation builds upon previous work \citep{SAT, LCM} in segmenting text into semantically coherent sentences. Specifically, we first utilize a pre-trained segmentation model to divide the response into sentences. We apply differential operations to reward
outputs at the start and end positions of each sentence to model the rewards of sentences. To facilitate training using readily available response-level preference labels, we employ an attention mechanism to aggregate the individual sentence rewards into a single reward score for the entire response. Finally, the sentence-level reward model is trained using the Bradley-Terry (BT) model \citep[BT,][]{bt} on response-level preference data. 
We conduct experiments to evaluate the performance of the sentence-level reward model. To evaluate the ability of the reward modeling, we test it on RewardBench~\citep{rewardbench}, where the sentence-level reward model outperforms the response-level reward model by 2.7\% on average. To evaluate the improvement in LLMs alignment with the sentence-level reward model, we apply two commonly used alignment algorithms: Best-of-N sampling (BoN, ~\citet{BoN}) and RLHF. On AlpacaEval~\citep{alpaca_eval}, our method surpasses the existing baseline approaches.

\section{Preliminaries}
\paragraph{Notations}
Given a prompt $x$ (consisting of several tokens), the LLMs $\pi_\theta(\cdot|x)$ generate a response $y = \left(a_1, a_2, \cdots, a_T\right)$, where $a_i$ represents a token in the vocabulary set $\mathcal{V}$ and $T$ represents the number of tokens.
We consider language generation as a Markov Decision Process (MDP), defined by a tuple $(S, A, \mathcal{P}, \mathcal{R}, T_{\mathrm{max}}, \gamma)$~\citep{rl_intro}. The prompt $x$ serves as the initial state $s_0$, and the action space $A$ is the vocabulary set $\mathcal{V}$. At time step $t$, an action $a_t \in A$ is generated by sampling from the policy $\pi_\theta(\cdot | s_t)$. The state $s_t \in S$ is determined by the transition function $\mathcal{P} : S\times A \rightarrow S$, typically by appending $a_{t-1}$ to $s_{t-1}$, resulting in $s_t = \left(s_{t-1}, a_{t-1}\right) = \left(x, a_1, \cdots, a_{t-1}\right)$. The reward function $\mathcal{R}: S \times A \rightarrow \mathbb{R}$ assigns a reward value $r(s_t, a_t)$ to each state-action pair $(s_t, a_t)$. The MDP terminates when $t$ reaches $T_{\max}$ or a special token (i.e., \texttt{<EOS>}) is sampled. $\gamma \in \left[0,1\right]$ is a discount factor.

\paragraph{Response-level Reward Modeling}
Response-level reward modeling is a widely used approach in RLHF~\citep{BoN,instruct-gpt,llama}, where a reward value is assigned to a complete response $(x, y)$. Formally, for a given $(s_t, a_t)$, its reward is defined as:
\begin{equation*}
    r(s_t, a_t) = 
    \begin{cases}
        r_\phi(x, y), & \text{if } t = T_{\mathrm{max}}\text{ or }a_t == \texttt{<EOS>}\\
        0, & \text{otherwise}
    \end{cases}
\end{equation*}
Here $r_\phi(x, y)$ denotes a parameterized reward model by $\phi$. To train it, we optimize the following BT loss: 
\begin{equation}
\label{bt_loss}
        \mathcal{L}(\phi) = -\mathbb{E}_{(x, y^w, y^l)\sim D_{\text{pref}}}\Bigl[ \sigma\left(r_\phi(x,y^w) - r_\phi(x,y^l)\right) \Bigr]
\end{equation}
where $\sigma(z) = 1/\left(1 + e^{-z}\right)$ is the sigmoid function, $D_{\text{pref}} = \{(x, y^w, y^l)\}$ denotes a preference dataset composed of prompt $x$, preferred response $y^w$ and less preferred response $y^l$.
\paragraph{Alignment with Reinforcement Learning}
Given a learned reward model $r_\phi(x, y)$ and a dataset $D_{\text{prompt}} = \{x^i\}_{i=1}^M$ containing $M$ prompts, 
we can employ reinforcement learning to fine-tune the LLMs $\pi_\theta(\cdot|x)$. Specifically, the objective can be formulated as follows:
\begin{equation}
\label{rlhf_obj}
\begin{split}
    \mathcal{L}(\theta) = & -\mathbb{E}_{x\sim D_{\text{prompt}}, y\sim \pi_\theta(\cdot|x)}\Bigl[r_\phi(x, y)
    - \beta D_{\mathrm{KL}}\left(\pi_\theta(y|x)||\pi_{\mathrm{ref}}(y|x)\right)\Bigr]
\end{split}
\end{equation}
where $\pi_{\mathrm{ref}}(\cdot|x)$ denotes the language model obtained via Supervised Fine-Tuning (SFT),
$D_{\mathrm{KL}}$ represents the KL divergence~\citep{KL}, serving as a regularization term that mitigates reward hacking~\citep{rm_scaling, odin} by constraining the deviation from the SFT policy. The hyperparameter $\beta$ controls the strength of the regularization term. We can optimize the aforementioned objective by RL algorithms~\citep{PPO, remax, reinforce++}.
\section{Method}
\label{method}
In this section, we detail the framework for training and utilizing sentence-level reward models. Section~\ref{train_srm} outlines the training process using preference datasets, and
Section~\ref{alignment_with_srm} describes how the trained model is used for LLM alignment. A workflow is illustrated in \cref{Framework}. 
\subsection{Sentence-level Reward Training}
\paragraph{Segmenting Responses into Sentences}
\label{split}
We begin by segmenting responses into sentences. Initial exploration considered rule-based methods \citep{moses,spacy}, which rely on fixed punctuation to mark boundaries. However, this approach faces challenges related to language specificity, requiring language identification, and difficulties with specialized content (e.g., math, code). For instance, dot characters (`.') may denote decimals or function calls, not just sentence endings.

Our segmentation method is based on SaT, a SOTA text segmentation model, and incorporates a rule-based segment approach for long sentences in SaT outputs. A detailed description is provided in \cref{app:seg}. Sentences obtained from segmentation operations are appended with a special boundary token (\texttt{<END>}). Only response sequences ($y^w, y^l$) are segmented; the prompt ($x$) is left unprocessed as it serves as an initial state in RLHF and is not directly optimized.

For a response token sequence $(x, a_1, \dots, a_T)$, segmentation yields a sequence of sentence chunks $(x, c_1, \dots, c_{n_c})$. Each chunk $c_i = (a_{b_{c_i}}, \dots, a_{d_{c_i}})$ consists of tokens from the original response, where $b_{c_i}$ and $d_{c_i}$ are the start and end token indices, and $n_c$ is the number of sentences.

\paragraph{Reward Modeling at the Sentence Level}
\label{method:aggregate}
We employ $r_\phi$ to project hidden states from the LLM backbone to a scalar reward. A naive approach to model the reward of sentence $c_i$ would use $r_\phi$'s output at the sentence-ending token ($a_{d_{c_i}}$). However, due to the autoregressive nature of LLMs, the hidden state at $a_{d_{c_i}}$ represents the entire preceding subsequence $(x, c_1, \dots, c_i)$, not just $c_i$. Thus, the reward obtained is for the subsequence $(x, c_1, \dots, c_i)$, creating a discrepancy with our objective of scoring $c_i$ alone $(i \geq 1)$. We therefore define the differential reward for $c_i$ as:
\begin{equation}
\label{eq:diff_reward}
\hat{r}(c_i) = r_\phi(x, c_1, \dots, c_i) - r_\phi(x, c_1, \dots, c_{i-1}). 
\end{equation}
A direct interpretation of $c_i$'s reward is the change in the subsequence reward caused by generating $c_i$. The similar approach is used in ~\citep{fine-rlhf} to detect sentence toxicity.
\paragraph{Aggregating into Response-Level Reward}
\label{train_srm}
To train with response-level preference labels, we need to aggregate sentence rewards into a response reward. 
We define the response reward as a weighted sum of sentence rewards:
\begin{equation}
    \label{r_phi}
        r_\phi(x, y) = \sum_{i=1}^{n_c} w_i \hat{r}(c_i),
\end{equation}
where  $\hat{r}(c_i)$ is given by Equation \eqref{eq:diff_reward},  $w_i$ determines the contribution of each sentence to the overall response. 
When no additional task-specific information is available, equal contribution is a natural choice. This approach is adopted by \citet{preference-grounded} to aggregate token-level rewards into response-level rewards, and similarly employed by \citet{IRCR} to distribute trajectory rewards into single steps. However, in the form we have defined, assuming each sentence contributes equally (i.e. $w_{n_c} = \cdots w_i \cdots = w_1$), we have $r(x, y) = w_{1}\left(r_\phi(x, c_1, \cdots, c_{n_c}) - r_\phi(x)\right)$, which degenerates into a response-level model. (\(r_\phi(x)\) will be eliminated when computing the BT loss, having no impact on training process)

Our basic motivation to design $w_i$ is that when generating a sentence $c_i$, if the subsequence and the full response are semantically closer, we should assign a higher weight to the sentence $c_i$. To achieve this, we introduce an attention mechanism to make $w_i$ a learnable parameter. We add two heads $q_\phi$ and $k_\phi$ besides $r_\phi$ and express $w_i$ as following:
\begin{equation*}
    w_i = \text{SoftMax}(q_{n_c}, k_i) = \frac{\exp(q_{n_c}^T k_i)}{\sum_{i=1}^{n_c} \exp(q_{n_c}^T k_i)}.
\end{equation*}
Here, \text{SoftMax} denotes the softmax operator, $q_{n_c} = q_\phi(x, c_1, c_2,\cdots,c_{n_c})$ represents a response-level query and $k_i = k_\phi(x, c_1, \dots, c_i)$ represents a key for subsequence $(x, c_1, \dots, c_i)$. Moreover, to effectively model the positional relationships across different sentences, we integrate Rotary Positional Embeddings (RoPE,~\citep{ROPE}) into the $q_{n_c}$ and $k_i$. A detailed description of the implementation is presented in \cref{rope_agg}.

Above all, we have presented an attention-based method for aggregating sentence rewards into response rewards. Subsequently, we can use BT loss~\eqref{bt_loss} to train the sentence-level reward model.
\subsection{Reinforcement Learning with Sentence-level Reward Model}
\label{alignment_with_srm}
The sentence-level reward model provides greater flexibility in aligning LLMs. The response reward derived by the aggregate function makes it compatible with existing response-level alignment methods, while utilizing sentence-level rewards allows for fine-grained optimization. We propose a method based on Reinforce++~\citep{reinforce++} for fine-grained LLM optimization.
\paragraph{Assigning Rewards to Sentence Boundary Tokens}
To utilize sentence-level rewards for policy optimization, we need to assign them to corresponding tokens in the LLM's output. Since LLMs do not output the special boundary token introduced in the RM training, we need an alternative method to identify sentence boundary tokens in sampled responses.

We propose a token-level matching method to address this issue. This method leverages the character offset mapping provided by a pre-trained BPE tokenizer \citep{bpe_src,bpe_tokenizer}, which links each token to its corresponding character range in the original response string. By comparing the sentence boundary character positions (determined by the segmenter) with these token character ranges, we identify the token whose range contains or immediately follows a boundary character. This identified token is designated as the sentence boundary token for receiving the sentence-level reward during policy optimization. Further details are in \cref{split_position}.

\paragraph{Policy Optimization with Sentence-level Reward}
We follow the policy optimization objective of Reinforce++ to solve objective~\ref{rlhf_obj}.
\begin{equation}
    \label{rpp_loss}
    \begin{split}
    \mathcal{L}^{\text{CLIP}}(\theta) &= -\mathop{\mathbb{E}}_{(s_t, a_t)\sim \pi_\theta}\Bigl[\text{min}\Bigl(r_t(\theta)\hat{A}(s_t, a_t), \text{clip}(r_t(\theta), 1 - \epsilon, 1 + \epsilon)\hat{A}(s_t, a_t)\Bigr)\Bigr].
\end{split}
\end{equation}
Indeed, this is the loss function of PPO~\cite{PPO}. $r_t(\theta) = \pi_\theta(a_t|s_t)/\pi^\text{old}_\theta(a_t|s_t)$ represents the ratio between the new policy $\pi_\theta(a_t|s_t)$ and old policy $\pi^{\text{old}}(a_t|s_t)$, the $\text{clip}$ function restricts the ratio $r_t(\theta)$ in $\left[1-\epsilon, 1+\epsilon\right]$, and $\hat{A}((s_t, a_t)$ denotes the advantage function, which is estimated using the following equation.
\begin{equation}
\label{adv_rpp}
    \hat{A}(s_t, a_t) = r(s_T, a_T) - \beta\sum_{i = t}^{T}\log\frac{\pi_\theta^\text{old}(a_i|s_i)}{\pi_\text{ref}(a_i|s_i)}.
\end{equation}
Only the final token $a_T$ receives the reward signal from the response-level reward model. We modify the estimation of the advantage function using the sentence-level reward scores.
\begin{equation}
    \label{srm_rpp_adv}
    \begin{split}
      \hat{A}(s_t, a_t) &= \sum_{i = t}^{T}\mathbb{I}_{\{i = c_d\}}w_ir_\phi(s_i, a_i) - \beta\sum_{i = t}^{T}\log\frac{\pi_\theta^\text{old}(a_i|s_i)}{\pi_\text{ref}(a_i|s_i)}.
    \end{split}
\end{equation}
Here, the indicator function $\mathbb{I}_{\{i = c_d\}}$ is used to identify if a token $a_i$ is a sentence boundary token.

\section{Related Work}
\subsection{Reward Learning from Human Preference}
\label{related work:reward learning}
In reinforcement learning, learning reward models from human preference is a common approach to designing reward functions.~\citet{drlhf} introduce the idea of learning a step-wise reward model from trajectory-level preference labels in deep reinforcement learning, and PT~\citep{pt} employs a transformer-based architecture to construct a non-Markovian reward model. 

With the widespread adoption of RLHF for LLM alignment, significant research has focused on developing effective reward models for LLMs. These efforts can be broadly categorized into discriminative RMs and generative RMs \citep{skywork-rm}. Discriminative RMs are currently widely utilized in post-training stages of LLMs~\citep{qwen2.5math, mathshepherd}, which project hidden states from backbone into reward scores with a scalar head, and are optimized through BT loss~\citep{bt}. To improve the performance of discriminative RMs, some studies have aimed at enhancing the quality of preference data~\citep{ultrafeedback, internlm2} or optimizing the model structure~\citep{moe_rm, odin}. Despite the popularity of discriminative RMs, generative RMs offer an alternative by leveraging the generative capabilities of the LLMs to directly evaluate the generated responses~\citep{llm-as-judge, generative_rms}. The training process of generative reward models typically aligns with language modeling, which allows it to improve the evaluation quality through prompt design and Chain-of-Thought \citep[CoT,][]{cot} techniques. 
For instance, GenRM~\citep{GenRM} combines the sequence to be evaluated with the question "\texttt{Is the answer right? (yes/no)}" as a prompt, using the probability of token \texttt{yes} as the reward score. 
Our work should be categorized as a discriminative reward model.
\subsection{Fine-grained Reward Model}
We consider the granularity of reward modeling. A common approach is to assign a reward score to an entire response. This coarse-grained method simplifies the reward modeling, but as discussed in \cref{intro}, it can lead to sparse reward challenges during RL training. To address the issue, many studies have explored assigning a reward score to a token, leading to fine-grained models. For instance, ~\citet{preference-grounded} directly sum the reward values of every token to compute the response reward and train it with the BT model. Moreover, ~\citet{densefree} use the attention weights from the trained response-level reward model to distribute response-level rewards to tokens. ~\citet{RTO} and ~\citet{r2q} train a policy model with DPO~\citep{DPO} to derive token-level rewards. In addition, ~\citet{tlcr} and ~\citet{drlc} leverage advanced AI systems to annotate token-level preferences.~\citet{seq2seq_rm} train a language model to modify sequences and construct discrete token rewards based on whether tokens are rewritten. 

Our work focuses on assigning reward scores to sentences. The most related approach is the Process Reward Model (PRM,~\citet{PRM}), which assigns rewards to steps in the problem-solving process and is often used to enhance the reasoning capabilities of LLMs. However, constructing PRMs typically requires manual annotation~\citep{PRM} or tree search~\citep{mathshepherd, omegaPRM} to label the correctness of each step. These approaches are both resource-intensive and challenging, especially for tasks where step correctness is ambiguous. In contrast, our work relies on the response-level preference labels.
\label{related work: fine-grained rm}

\section{Experiments}
\label{exp}
\subsection{Settings}
\paragraph{Dataset}
For reward model training, we use the binarized version\footnote{\url{https://huggingface.co/datasets/HuggingFaceH4/ultrafeedback_binarized}} of Ultrafeedback~\citep{ultrafeedback}, comprising 64k prompts and 128k responses with human preference labels. For the subsequent alignment experiments, we utilize the HH-Golden\footnote{\url{https://huggingface.co/datasets/Unified-Language-Model-Alignment/Anthropic_HH_Golden}} dataset. This dataset is a processed version of Anthropic's Helpful and Harmless (HH) dataset~\citep{hh-rlhf}, comprising 42.5k high-quality prompts and corresponding responses. Specifically, following the methodology of DeepSpeed-Chat\footnote{\url{https://github.com/deepspeedai/DeepSpeedExamples/tree/master/applications/DeepSpeed-Chat}}~\citep{ds-chat}, the HH-Golden dataset is split into two equally sized subsets (5:5 ratio) used for Supervised Fine-Tuning (SFT) and RLHF, respectively. In the RL phase, only the prompts from the allocated subset are used for response generation; the original responses from this subset are not utilized for training.
\paragraph{Models and Training}
We use \texttt{sat-3l}~\citep{SAT} as the segmenter. For reward model training, we use Llama3.1-8b~\cite{llama3} as the base model. For the RL experiments, we first perform SFT on Llama3.2-3b, and then used the SFT model as the initial model. In addition, to verify the scaling effect of our method, we also conduct RL experiments using an open-source SFT model Llama3.1-Tulu-3-8B-SFT. For the BoN experiments, we generate responses using Llama-3.1-8B-Instruct. The framework for reward model training is based on DeepSpeed-Chat, and the RL training framework is built on Open-RLHF\footnote{\url{https://github.com/OpenRLHF/OpenRLHF}}. More detailed experimental parameter configurations are provided in \cref{exp_config}.

\paragraph{Evaluation}
We evaluate the performance of our reward model on RewardBench~\citep{rewardbench}, a preference dataset comprising 2,985 samples. The dataset is further categorized by task: Chat, Chat-Hard, Safety, and Reasoning. Reward model performance is measured by its accuracy in predicting the ground truth chosen/rejected labels. For our sentence-level reward model, evaluation on RewardBench uses the aggregated response reward derived from the aggregation function (\cref{r_phi}).

To evaluate the alignment performance of LLMs fine-tuned using reward models, we utilize the AlpacaEval~\citep{alpaca_eval} and Arena-Hard~\citep{arena} benchmarks. AlpacaEval comprises 805 prompts, while Arena-Hard provides 500 challenging prompts. Responses generated by the evaluated LLMs are compared by a judge model. We use DeepSeek-V3~\citep{Deepseek-v3} as the judge model and report the win rate of LLMs fine-tuned against the initial SFT model. In addition, we conduct Best-of-N (BoN, \citep{BoN}), a common inference-time alignment technique. BoN results are also evaluated on AlpacaEval.

\paragraph{Baselines}
For comparison, we employ two standard reward models as baselines: a response-level reward model \citep{BoN} and a token-level reward model \citep{preference-grounded}. For brevity, we refer to the response-level and token-level reward models as \textbf{Response} and \textbf{Token}, respectively. For the RL experiments, we define two additional baselines based on different reward signal granularities:
\begin{itemize}
\item Response to Sentence (\textbf{Res2Sent}): This baseline adapts a trained response-level reward model to provide sentence-level signals. The reward for a sentence is computed as the difference between the response-level model's score at the sentence end position and its score at the start position. These sentence rewards are then used for RL training following the same fine-grained approach as our method (\cref{srm_rpp_adv}).
\item Sentence to Response (\textbf{Sent2Res}): This baseline uses our trained sentence-level reward model to provide a single response-level score. Specifically, the reward for a response is the aggregated score from \cref{r_phi}. This response score is then used for standard RL training, following the approach for response-level models (\cref{adv_rpp}).
\end{itemize}
\begin{figure*}[htbp]
    \centering
    \includegraphics[width=\linewidth, trim = 0.1cm 0.1cm 0.2cm 0.1cm, clip]{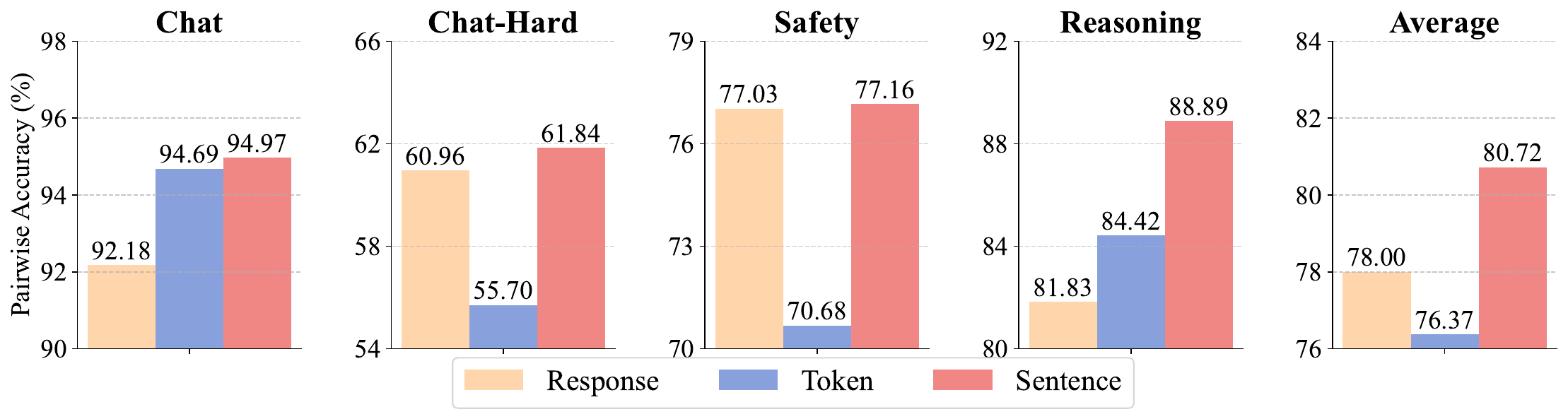}
    \caption {\textbf{The prediction accuracy of the reward models on RewardBench}. Our sentence level reward model outperforms the token-level and response-level reward models on all subtasks.}
    \label{fig:reward_bench_comparison_bon}
\end{figure*}
\vspace{-15pt}
\begin{table}[htbp]
\centering 
\small
\setlength{\abovecaptionskip}{2pt} 
\setlength{\belowcaptionskip}{-5pt} %
\renewcommand{\arraystretch}{1.1}
\setlength{\tabcolsep}{5pt} 
\makebox[\textwidth][c]{
\begin{tabular}{l|ccc|cc|ccc|cc}
\toprule
\multirow{2}{*}{\textbf{Method}} & \multicolumn{5}{c|}{\textbf{Llama3.2-3B-SFT}} & \multicolumn{5}{c}{\textbf{Llama-3.1-Tulu-3-8B-SFT}} \\
\cmidrule(lr){2-6} \cmidrule(lr){7-11}
& \multicolumn{3}{c|}{\textbf{AlpacaEval}} & \multicolumn{2}{c|}{\textbf{Arena-Hard}} & \multicolumn{3}{c|}{\textbf{AlpacaEval}} & \multicolumn{2}{c}{\textbf{Arena-Hard}} \\
& LC (\%) & WR (\%) & Len. & WR (\%) & Len. & LC (\%) & WR (\%) & Len. & WR (\%) & Len. \\
\midrule
Response       & 55.97 & 71.55 & 467 & 74.90 & 481 & 59.73 & 69.25 & 1587 & 58.70 & 2044 \\
Token          & 55.06 & 70.06 & 2190 & 74.70 & 2016 & 60.88 & 73.60 & 2105 & \textbf{63.80} & 2278 \\
Res2Sent       & 57.90 & 54.91 & 219  & 48.80 & 220 & 65.03 & 68.76 & 1437 & 61.40 & 1936 \\
Sent2Res       & 60.32 & 77.27 & 623  & 74.40 & 583 & 63.41 & 70.87 & 1550 & 58.70 & 2063 \\
\textbf{Sentence}(ours)  & \textbf{60.32} & \textbf{78.39} & 558 & \textbf{76.00} & 638 & \textbf{71.52} & \textbf{75.65} & 1447 & 63.00 & 2021 \\
\bottomrule
\end{tabular}
} 
\caption{The win rates(WR) and Length-Control win rates(LC) of different models compared to SFT on AlpacaEval2 and Arena-Hard benchmarks. Results demonstrate the absolute superiority of our proposed method over alternative approaches. Especially, our method achieves a favorable trade-off between generation quality and response length.}
\label{tab:alpaca}
\end{table}
\subsection{Main Results}
\cref{fig:reward_bench_comparison_bon} presents the performance of our sentence-level reward model and baseline models on RewardBench. Across the four subtasks and in overall average performance, the sentence-level reward model consistently outperforms the token-level and response-level reward models. On average, our sentence-level model surpasses the second-best baseline (response-level) by 2.7\%. Particularly, on the reasoning task, our sentence-level reward model significantly outperforms other reward models, leading the response-level and token-level models by 7.1\% and 4.4\%, respectively. Under identical conditions regarding the base model and training data, the above results demonstrate the improved generalization capability of our proposed sentence-level reward model to unseen preference datasets.

\cref{tab:alpaca} presents the performance of LLMs fine-tuned with various reward models on AlpacaEval and Arena-Hard. All win rates are reported relative to the initial SFT model. Based on these results, the following key observations can be drawn:

(1) \textbf{RL training can benefit from sentence-level reward signals}. For the same sentence-level reward model, training using direct sentence-level rewards yields higher performance compared to training with the aggregated response score (Sent2Res). And LLMs trained with sentence-level reward model consistently outperform those trained with the Response baseline. This indicates that incorporating sentence-level reward signals significantly enhances the effectiveness of RL training.

(2) \textbf{The sentence-level reward model exhibits greater robustness against length bias}. Both sentence-level and token-level reward models provide fine-grained signals for RL training. However, LLMs trained with token-level reward suffer severely length hacking, generating responses significantly longer than other methods and showing a clear degradation in LC win rates. Our sentence-level reward model guides the model to improve generation quality while maintaining controlled response length.

(3) \textbf{The response-level reward model struggles to provide accurate reward signals for sentences}. We applying similar differential operations for the response-level reward model to derive sentence-level rewards(Res2Sent). The overall performance of LLMs trained with Res2Sent lags behind those trained with the direct response reward. It's inferred response-level reward models provide inaccurate reward signals for sentences, making it difficult for RL training to benefit from it.

\begin{figure}[htbp]
    \centering
    \begin{subfigure}{0.31\textwidth}
        \centering
        \includegraphics[width=1.05\textwidth]{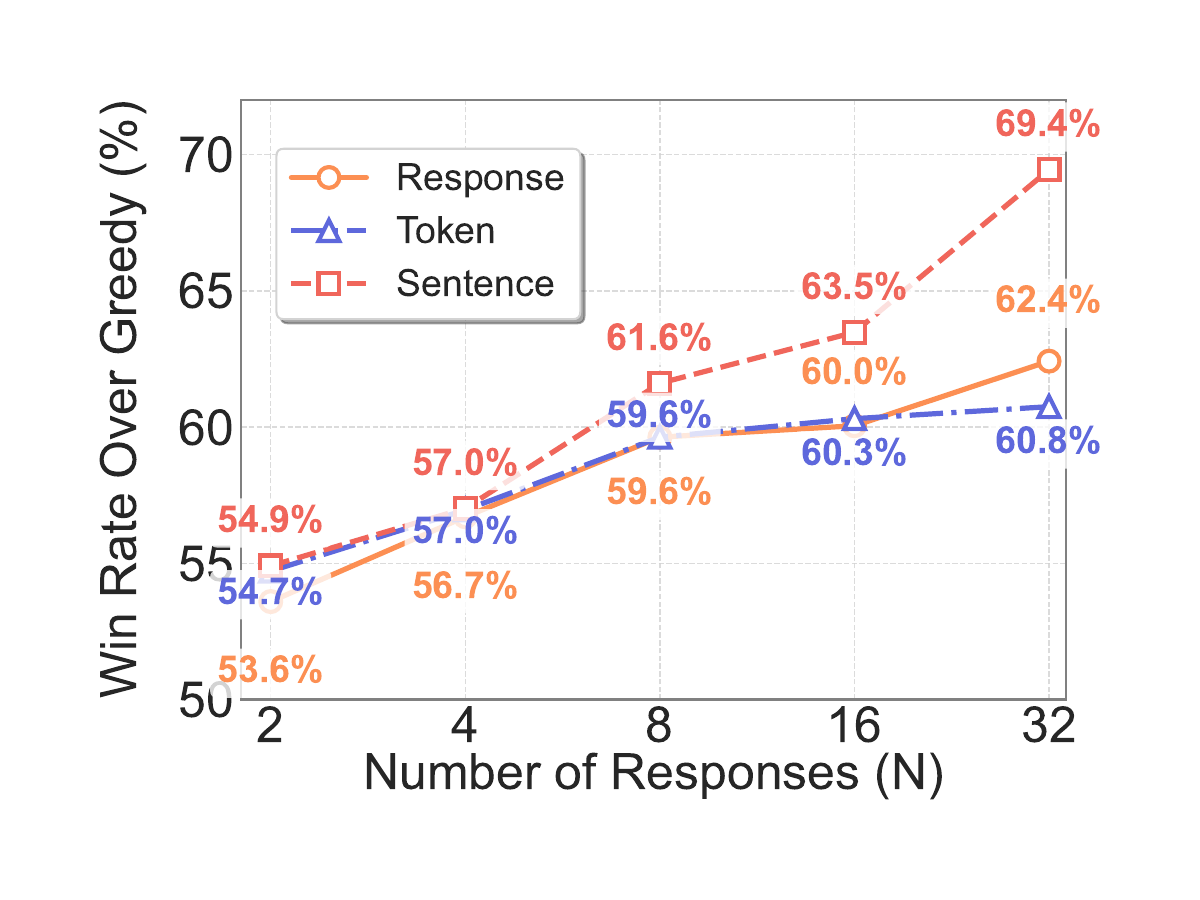}
        \caption{\textbf{BoN results on AlpacaEval}. We vary N from 2 to 32. Win rate is calculated by comparison against greedy decoding.The sentence-level reward model exhibited the best performance.}
        \label{fig:bon_alpaca}
    \end{subfigure}
    \hspace{0.02\textwidth}
    \begin{subfigure}{0.31\textwidth}
        \centering
        \includegraphics[width=1.05\textwidth]{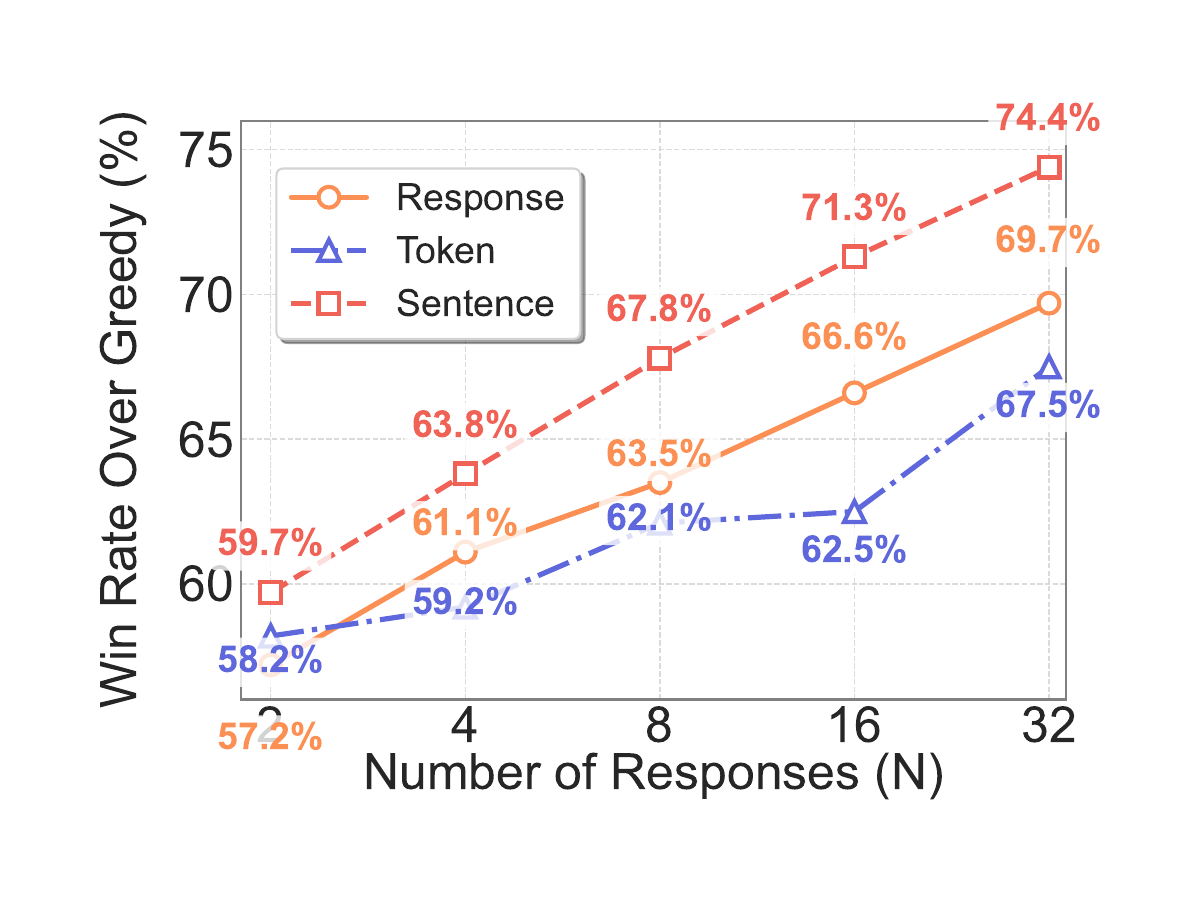}
        \caption{\textbf{BoN results on Arena-Hard}. We vary N from 2 to 32. Win rate is calculated by comparison against greedy decoding.The sentence-level reward model exhibited the best performance.}
        \label{fig:bon_arena}
    \end{subfigure}
    \hspace{0.02\textwidth}
    \begin{subfigure}{0.31\textwidth}
        \centering
        \includegraphics[width=0.9\textwidth]{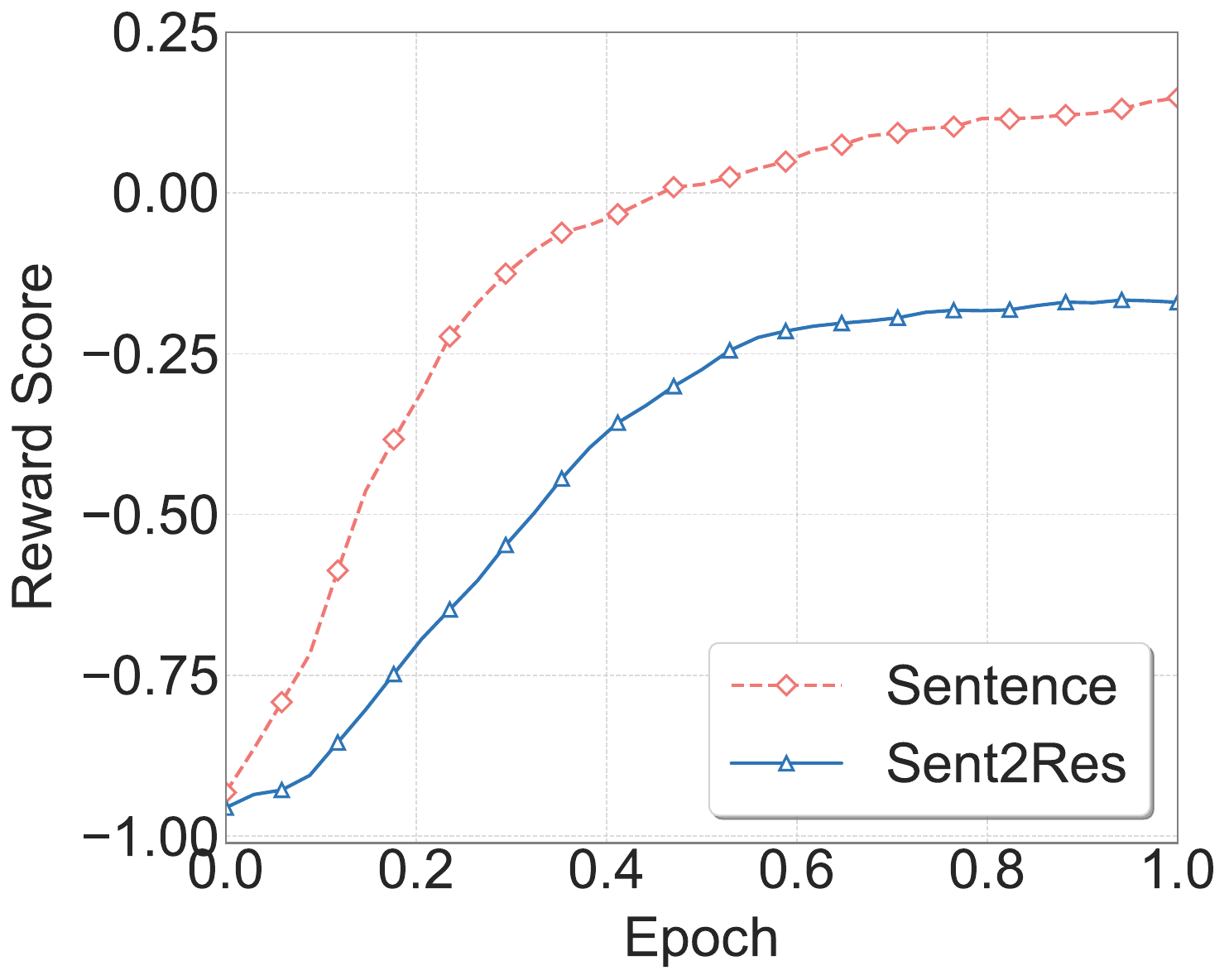}
        \caption{\textbf{Reward curves during RL training}. Directly using sentence-level reward signals exhibits higher training efficiency than aggregating them into response-level signals.}
        \label{fig:reward_compare}
    \end{subfigure}
\end{figure}

\cref{fig:bon_alpaca} and \cref{fig:bon_arena} illustrate the performance of various reward models in the context of BoN alignment. We vary the value of $N$ and report the win rate against greedy decoding. The sentence-level reward model consistently outperforms both the token-level and response-level reward models across different settings. Notably, when $N=32$, our method achieves a 7\% performance improvement over the baselines on AlpacalEval. These results highlight the potential of utilizing sentence-level reward models to enhance inference-time alignment strategies
\setlength{\intextsep}{5pt}
\vspace{-5pt}
\vspace{-15pt}
\subsection{Training Efficiency Improvements}

We investigate whether sentence-level reward signals improve RL training efficiency. \cref{fig:reward_compare} presents RL training curves of Llama-3.1-Tulu3-8B-SFT comparing two approaches that utilize signals from the same sentence-level reward model: one uses the aggregated response score for coarse-grained training, and the other uses direct sentence-level rewards for fine-grained training. Performance is evaluated using the aggregated response score (\cref{r_phi}). As depicted in \cref{fig:reward_compare}, training with direct sentence-level rewards yields substantially higher evaluation scores within the same training duration. This indicates that fine-grained sentence-level reward signals facilitate more efficient sample utilization, leading to improved training efficiency.
\subsection{Reward visualization}

\begin{wrapfigure}[23]{r}{0.6\textwidth}
  \vspace{-45pt}
  \includegraphics[width=0.6\textwidth]{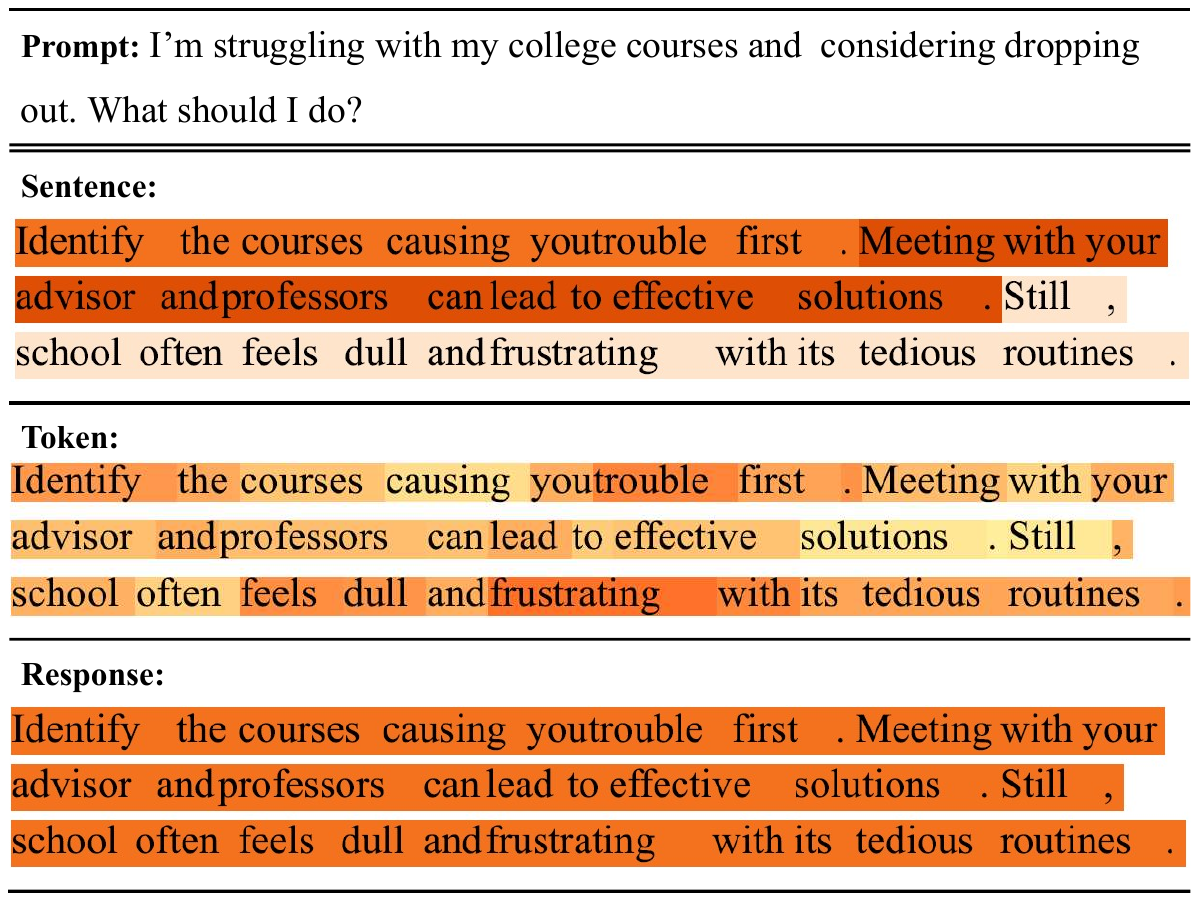}
  \captionsetup{aboveskip=-140pt} 
  \caption{\textbf{Reward distribution analysis} for different models on an educational advice response. Our sentence-level model (top) selectively rewards actionable suggestions (sentences 1-2, dark shading) while penalizing unconstructive complaints (sentence 3). Token-level rewards (middle) lack clear semantic interoperability and response-level scoring (bottom) provides uniform evaluation. Color intensity corresponds to reward magnitude.}
  \label{fig:demo}
\end{wrapfigure}

\cref{fig:demo} illustrates the reward values assigned by different reward models. Color intensity indicates reward magnitude, with darker shades representing higher values. As shown, the response-level reward model assigns a uniformly high score across the entire generated response, even for segments with relatively weak relevance to the prompt(e.g., the third sentence). In contrast, the sentence-level reward model demonstrates a more refined understanding of semantics. It can effectively assess the relevance of each sentence to the prompt, assigning higher rewards to strongly aligned content and penalizing sentences that diverge from the intended meaning. We believe this contributes to the sentence-level reward model's greater robustness against length bias. While the token-level reward model also provides fine-grained signals, our analysis suggests these signals often lack a robust and consistent correlation with the underlying semantic meaning, potentially leading to inaccurate reward signals.

\setlength{\belowcaptionskip}{-10pt}
\begin{figure}[htbp]
    \centering
    \begin{minipage}[b]{0.48\linewidth}
        \centering
        \includegraphics[width=1.05\linewidth]{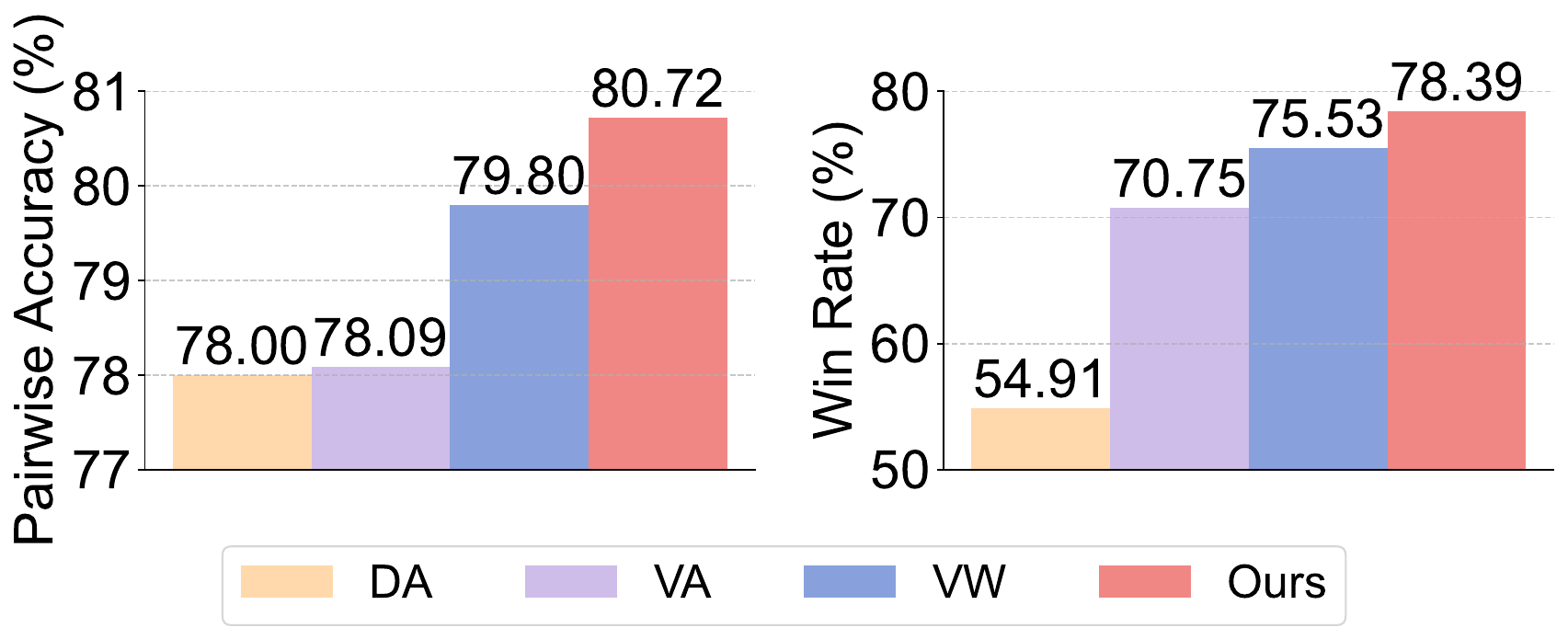}
        \caption{Comparison using different aggragation functions to train reward model. \textbf{Left}: The accuracy of trained reward models on RewardBench. \textbf{Right}: The win rate over SFT on AlpacaEval. The proposed aggregation function demonstrates the best performance.}
        \label{fig:agg_fun}
    \end{minipage}
    \hfill
    \begin{minipage}[b]{0.48\linewidth}
        \centering
        \includegraphics[width=1.05\linewidth]{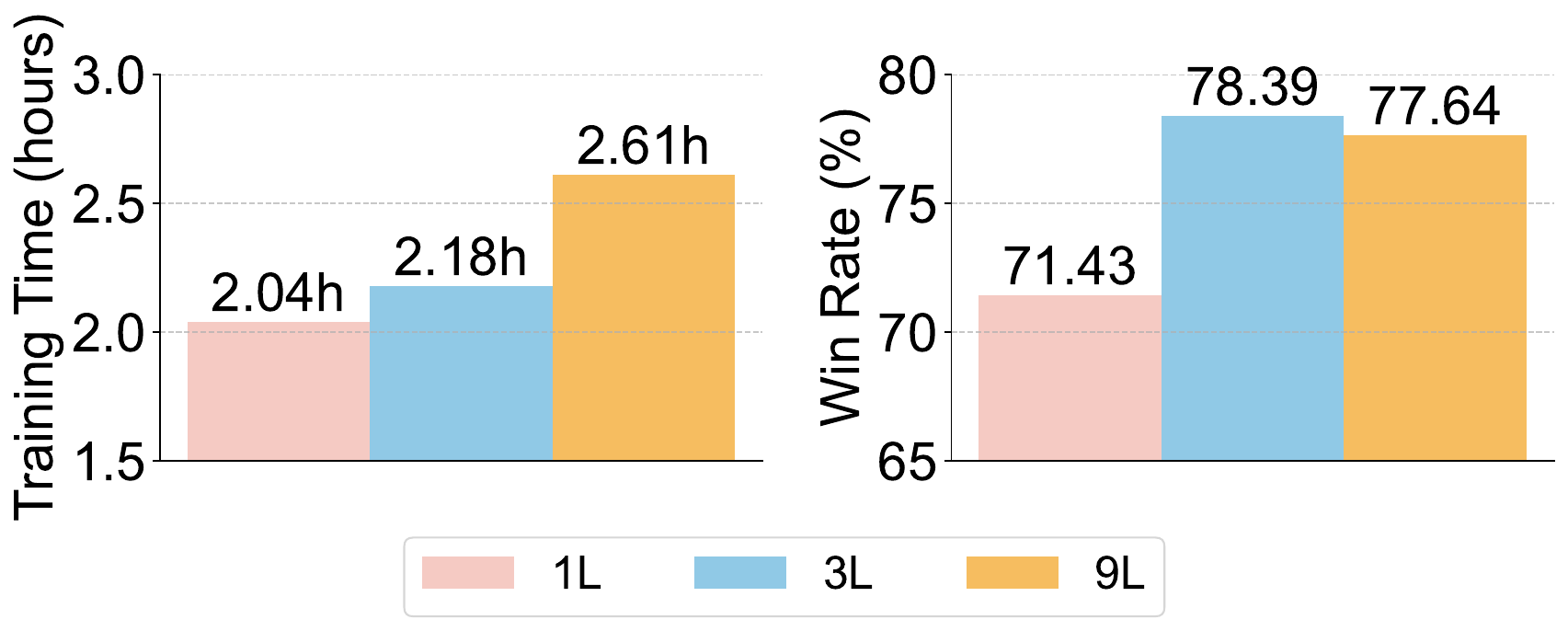}
        \caption{Comparison using different SaT models as sentence segmenter. \textbf{Left}: Training time of RLHF. \textbf{Right }: Win rates over SFT model on AlpacaEval. \texttt{sat-3l} strikes a favorable trade-off between training time and overall alignment performance.}
    \label{fig:ab_sat}
    \end{minipage}
\end{figure}
\subsection{Ablation Studies}
\paragraph{The Impact of the Aggregation Functions}
\label{agg_fn}
Our proposed method utilizes differential operations for sentence reward computation and weighted summation for response-level aggregation. To evaluate the contribution of these components, we compare our approach to variants using alternative designs for sentence reward calculation and aggregation. We explore the following combinations:
\begin{itemize}
    \item \textbf{VW}: Using the \textbf{v}alue head's raw output as sentence rewards and aggregating them with \textbf{w}eighted summation.
    \item \textbf{VA}: Using the \textbf{v}alue head's raw output as sentence rewards and aggregating them with \textbf{a}verage rewards of all sentences.
    \item \textbf{DA}: Using the \textbf{d}ifferential operations to get sentence rewards and aggregating them with \textbf{a}verage rewards of all sentences.
\end{itemize}

Specific formulations for these aggregation functions are provided in \cref{form_agg}. Performance is reported in \cref{fig:agg_fun}, measured by RewardBench accuracy and AlpacaEval win rate. As shown in \cref{fig:agg_fun}, variants without differential operations for sentence rewards (VW, VA) and without weighted summation for aggregation (VA, DA) generally perform worse. Specifically, the DA variant performs worst (both reward modeling and alignment). These results highlights the effectiveness of the proposed differential operation and weighted summation in accurately modeling sentence values and aggregating them into a meaningful response score.
\paragraph{The Impact of the Segment Models}

In our main experiments, we used the \texttt{sat-3l} model as the sentence segmenter. To investigate the impact of using different SaT model variants, we evaluated the performance of three models: \texttt{sat-1l}, \texttt{sat-3l}, and \texttt{sat-9l}. Win rates on AlpacaEval and corresponding RLHF training times are presented in \cref{fig:ab_sat}. Figure \cref{fig:ab_sat} shows that the \texttt{sat-9l} model performs the worst, likely due to its limited segmentation capability. While \texttt{sat-9l} achieves similar alignment performance to \texttt{sat-3l}, it results in a notable 19.7\% increase in training time. Considering the trade-off between alignment performance and training efficiency, `sat-3l` is selected for our main experiments.

\section{Summary}
\label{summary}
In this study, we introduce a novel sentence-level reward model designed to provide finer-grained reward signals for RLHF. We propose a differential-based method for sentence-level reward modeling and employ an attention mechanism for aggregation, enabling training with response-level preference labels and the BT loss. Additionally, we present a fine-grained optimization framework utilizing the sentence-level reward model. Experimental results demonstrate its superior performance in both reward modeling and LLMs alignment. However, due to limited resources and time, experiments were restricted to Reinforce++\citep{reinforce++} and 8b models. Future work will involve exploring extensions to larger base models and more diverse datasets, and integrating the sentence-level RM with additional RL algorithms.
\newpage
\bibliography{reference}
\newpage
\newpage
\appendix
\section{Method Details}
\subsection{Pseudo-code for Reward and Policy Training}
The pseudo-code of sentence level RM and LLMs training is presented in \cref{alg:reward_policy_training}. 
\begin{algorithm}[H]
\caption{Reward and Policy Training}
\label{alg:reward_policy_training}
\begin{algorithmic}
\STATE \textbf{Input:} 
Preference dataset $ D_{\text{pref}} = \{(x, y^w, y^l)\}_{i = 1}^{N}$ for reward model training, a prompt dataset $ D_{\text{prompt}} = \{x^i\}_{i = 1}^{M} $ for policy training, a fine-tuned SFT model,  the number of iterations $ M_{\text{reward}} $ for reward model training, the number of iterations $M_{\text{policy}}$ for policy training,  KL divergence coefficient $\beta$.

\STATE \textcolor{gray}{// Training the sentence reward model}
\STATE Preprocess the preference dataset \(D_{\text{pref}} \) using the SaT model by inserting the <END> separator at the end of each segmented sentence and concatenating the sentences into complete responses. The processed dataset is denoted as \( D_{\text{END}} \).
\FOR{$\mathrm{iter} \in  \{ 1, \ldots, M_{\mathrm{reward}}\}$}
\STATE Sample a mini-batch $\mathcal{B} = \{(x^i, y^{w,i}, y^{l,i})\}_i \sim D_{\text{END}} $.
\STATE Compute rewards using \cref{r_phi}.
\STATE Optimize the reward model using \cref{bt_loss}.
\ENDFOR
\STATE

\STATE \textcolor{gray}{// Training the policy with the sentence reward}
\FOR{$\mathrm{iter} \in\{ 1, \ldots, M_{\text{policy}}\}$}
    \STATE Sample a mini-batch $\mathcal{B} = \{x^i\}_i \sim D_{\text{prompt}} $.
    \STATE Sample a response $y^i \sim \pi_{\theta}(\cdot|x^i) $ for each $x^i \in \mathcal{B}$.
    \STATE Process $y^i$  using the SaT model to insert <END> separators, resulting in $y^i_{\text{split}}$.
    \STATE Compute sentence-level rewards $r_{\phi}(x^i, y^i_{\text{split}})$ using the reward model.
    \STATE Map the rewards of $y^i_{\text{split}}$ to the corresponding tokens in $y^i$ to obtain $\hat{r}(c_i)$ as described in \ref{split_position}.
    \STATE Optimize the policy model using the sentence-level rewards $\hat{r}(c_i)$ with \cref{rpp_loss}.
\ENDFOR
\end{algorithmic}
\end{algorithm}

\subsection{ROPE for Attention-based Aggregate Function}
\label{rope_agg}
To effectively
model the positional relationships across different sentences, we integrate Rotary Positional Embeddings (RoPE, ~\citep{ROPE}) into the $q_{n_c}$ and $k_i$. We provide a detailed explanation of the integration method of RoPE. As one of the most widely adopted positional encoding methods in LLMs, RoPE is implemented by applying a position-dependent rotation matrix to the query ($q$) and key ($k$) vectors within the attention block. This rotation matrix encodes the positional information directly into the representations. If a token $x_i$ is located at position $i$ in the sequence, and its corresponding query $q_i$ and key $k_i$ have a dimensionality of $d$, then the rotation matrix $W^R_i$ to be applied is defined as follows:
\begin{equation*}
W^R_i = \left(\begin{array}{ccccccc}
\cos i \theta_0 & -\sin i \theta_0 & 0 & 0 & \cdots & 0 & 0 \\
\sin i \theta_0 & \cos i \theta_0 & 0 & 0 & \cdots & 0 & 0 \\
0 & 0 & \cos i \theta_1 & -\sin i \theta_1 & \cdots & 0 & 0 \\
0 & 0 & \sin i \theta_1 & \cos i \theta_1 & \cdots & 0 & 0 \\
\vdots & \vdots & \vdots & \vdots & \ddots & \vdots & \vdots \\
0 & 0 & 0 & 0 & \cdots & \cos i \theta_{d / 2-1} & -\sin i \theta_{d / 2-1} \\
0 & 0 & 0 & 0 & \cdots & \sin i \theta_{d / 2-1} & \cos i \theta_{d / 2-1}
\end{array}\right)
\end{equation*}
Here $\theta_t = 1000^{-\frac{2(t - 1)}{d}}, t = 1, 2, \cdots, d / 2$. The rotated $q^R_i = W^R_i q_i$ and $k^R_i = W^R_ik_i$ are then used to compute the dot product, yielding the corresponding attention weights. In our sentence-level reward model, two types of positional information are present: (1) the position of a sentence within the overall response, and (2) the position of sentence-boundary tokens within the input sequence. We apply the RoPE operation based on the sentence-level positional information.
\subsection{Assigning Rewards to Sentence Boundary Tokens}  
\label{split_position}  
Our sentence-level reward model operates on responses processed by the SaT model, which segments the text into sentences. After obtaining the segmentation results from SaT, we manually insert special \texttt{<END>} tokens to mark the boundaries between sentences. However, since the \texttt{<END>} tokens are not part of the policy's original output and are irrelevant to the optimization process, we need to identify the token positions immediately preceding each \texttt{<END>} token. These positions serve as the anchor points for reward assignments.  

A straightforward approach is to segment the response and tokenize each sentence separately, allowing us to identify the last token as the sentence boundary. We then concatenate the tokenized sequences of all sentences as the final tokenization result for the entire response. However, we observe that under BPE-style tokenizer~\citep{bpe_src,bpe_tokenizer}, this approach leads to mismatches—specifically, the concatenation of individually tokenized sentences does not always match the result of tokenizing the entire sequence directly. A key challenge arises from the tokenizer's behavior of merging multiple characters into a single token. For example, multiple consecutive spaces may be encoded as a single token.

To address this, we leverage the tokenizer's \texttt{offset\_mapping} feature, which maps each token back to its character-level position in the original text. We hope to construct a binary mask that aligns with the policy's token sequence. In this mask, \texttt{1} indicates a boundary position where rewards are assigned, while \texttt{0} indicates non-boundary positions with a zero reward.  

The implementation involves the following two main steps:  

\begin{enumerate}  
    \item \textbf{Tokenization and Boundary Identification}:  
    The policy's original response (\texttt{src\_text}) is tokenized using the tokenizer, with \texttt{offset\_mapping} enabled to track character-level positions. Simultaneously, the response with manually inserted \texttt{<END>} tokens (\texttt{split\_text}) is split by the \texttt{<END>} token to identify clause boundaries at the character level.  

    \item \textbf{Mask Generation}:  
    Using the \texttt{offset\_mapping}, we map the character-level boundary positions to their corresponding token indices in the policy's token sequence. A binary mask is then created, where \texttt{1} is assigned to tokens immediately preceding the identified boundaries, and \texttt{0} is assigned to all other tokens.  
\end{enumerate}  

This approach ensures that the reward computation is accurately aligned with the sentence boundaries, even when the tokenizer merges multiple characters into a single token.  

\subsection{Reward Model Training without Introducing the Special Token}
We introduce a special token to mark the end of a sentence, and the sentence-level reward model outputs the sentence's reward value at the special token's position. In fact, by using the method mentioned in \cref{split_position}, we can directly identify the sentence boundary token's position without introducing the special token. We train the sentence-level reward model with this approach, then we follow the same RL experiment configurations as ~\cref{exp} to evaluate the performance. We report the win rate of the sentence-level reward model(without special token) compared to baselines and our method (with special token) in \cref{tab:alpaca_wo_end_1}. It was observed that, compared to the response-level and token-level reward model, it achieves a higher win rate against SFT model, indicating that sentence-level modeling can offer a performance over the response-level reward modeling. However, when compared to the model with the special token (\texttt{<END>}), its win rate show a slight drop, suggesting that the approach with the special token leads to better sentence-level reward model training.

Additionally, we explore the impact of different aggregation functions on the performance of the reward model. We train the reward model using the aggregation functions described in \cref{agg_fn}, with the key distinction that the \texttt{<END>} token is excluded in this experiment using the method metioned in\cref{split_position}. Notably, since using the aggregation function \textbf{DA} during training causes it to degrade into a response-level reward model, we don't conduct experiments for this case. The trained reward models are then utilized for reinforcement learning. We evaluate the results on the AlpacaEval benchmark, computing the win rates by comparison with the Response. The detailed results are presented in \cref{tab:alpaca_wo_end_2}. The results reveals that the our proposed aggregation function (without \texttt{<END>}) achieves the best overall performance in terms of both win rate (WR) and length control win rate (LC). 


\begin{table}[!h]
  \centering
  \begin{minipage}{0.48\textwidth}
    \setlength{\tabcolsep}{6.0pt} 
    \small
    \renewcommand{\arraystretch}{1.} 
    \centering
    \begin{tabular}{l | c | c}
      \toprule
      \multirow{1}{*}{Method} & WR(\%) & LC(\%)\\
      \midrule
      Response & 71.55 & 55.97\\
      Token & 70.06 & 55.06 \\
      \textbf{Ours} (w.o. \texttt{<END>})& 74.66 & 60.31\\ 
      \textbf{Ours} (w. \texttt{<END>})& \textbf{78.39} & \textbf{60.32} \\
      \bottomrule
    \end{tabular}
    \caption{Training sentence-level reward model(w.o. \texttt{<END>}), we evaluate the Win Rate(WR) and Length Control Win Rate(LC) against SFT. The performance of Sentence (w.o. \texttt{<END>}) surpasses that of Response and Token but falls behind Sentence (w. \texttt{<END>}).}
    \label{tab:alpaca_wo_end_1}
  \end{minipage}
  \hfill
  \begin{minipage}{0.48\textwidth}
    \setlength{\tabcolsep}{6.0pt} 
    \small
    \renewcommand{\arraystretch}{1.} 
    \centering
    \begin{tabular}{l | c | c}
      \toprule
      \multirow{1}{*}{Method} & WR(\%) & LC(\%)\\
      \midrule
      VA & 73.91 & 57.44\\
      VW & \textbf{75.78} & 55.99 \\
      \textbf{Ours} (w.o. \texttt{<END>}) & 74.66 & \textbf{60.31}\\  
      \bottomrule
    \end{tabular}
    \caption{Using different aggregation functions (w.o. \texttt{<END>}), we evaluate the win rate (WR) and length control win rate (LC) compared to the Response. Our aggregation function achieves the best overall performance in terms of both WR and LC metrics.}
    \label{tab:alpaca_wo_end_2}
  \end{minipage}
\end{table}
\subsection{Strategy for Sentence Segmentation}
\label{app:seg}
SaT~\citep{SAT} provides a range of models for predicting sentence boundaries. We select the \texttt{sat-3l} model as our base segmenter, as it offers a favorable balance between segmentation quality and inference time. However, the output of the \texttt{sat-3l} segmenter may still include excessively long sentences. Following the approach in ~\cite{LCM}, we implement a supplementary step: we set a maximum character limit for a single sentence, and any segments exceeding this limit are re-segmented using a rule-based approach. The segmentation markers used are as follows: [\texttt{"..."},\texttt{"\string\n"},\texttt{"!"},\texttt{"?"},\texttt{";"},\texttt{":"},\texttt{"."},\texttt{","},\texttt{"\string\t"}].
\subsection{Early Explorations for Sentence-level Reward Model}
\begin{figure}[htbp]
    \centering
    \includegraphics[width=0.8\linewidth]{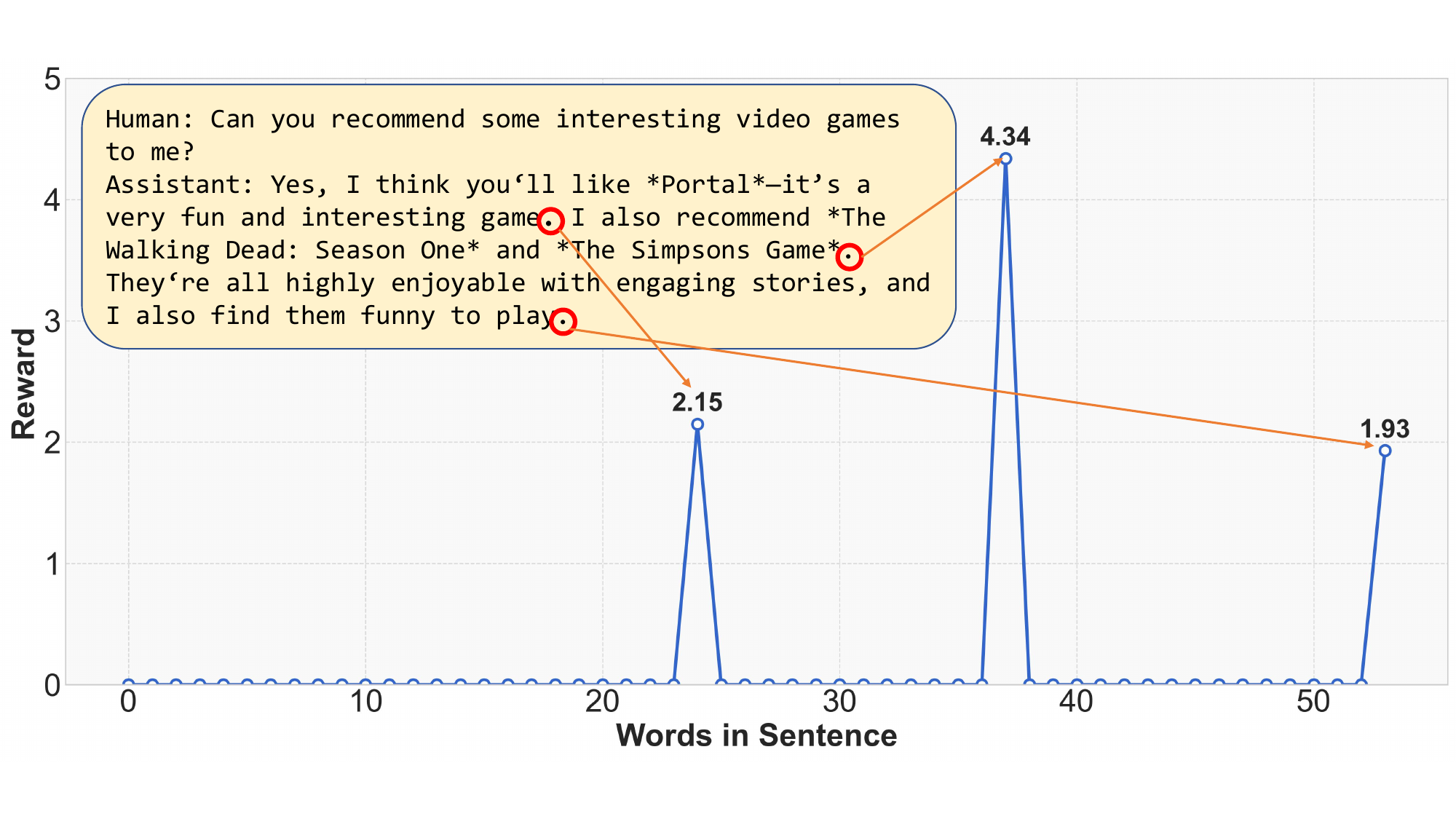}
    \caption{Reward distribution for token-level reward model trained with PT objective. It's observed that tokens at sentence boundaries exhibit significantly higher weighted reward values. }
    \label{fig:pt_trm}
\end{figure}

The idea of sentence-level reward model originates from our attempts to extend the PT~\citep{pt} approach to LLMs. Our initial exploration involves training a token-level reward model using a PT objective:
\begin{equation*}
    P(y^w > y^l|x) = \frac{\mathrm{exp}\left(\sum_{t = 1}^Tw_\phi(s^w_t, a^w_t)r_\phi(s^w_t, a^w_t)\right)}{\sum_{j\in\{w, l\}}\mathrm{exp}\left(\sum_{t = 1}^Tw_\phi(s^j_t, a^j_t)r_\phi(s^j_t, a^j_t)\right)}
\end{equation*}
where $w_\phi(s^j_t, a^j_t) = \frac{1}{T}\sum_{r = 1}^{T}\mathrm{SoftMax}(q_t^j, k_r^j)$. $q_t^j = q_\phi(s_t^j, a_t^j)$ and $k_t^j = k_\phi(s_t^j, a_t^j)$ are obtained from the corresponding projection layer.
While this token-level RM does not achieve superior performance compared to the response-level baseline, we make an interesting observation. Specifically, when analyzing the reward distribution of a PT-trained reward model, we find that tokens at sentence boundaries exhibit significantly higher weighted reward values, as is illustrated in \cref{fig:pt_trm}. This observation directly inspires us to attempt reward modeling at the sentence level.

\section{Experiments Configurations}
\subsection{Parameter Configurations}
\label{detail_exp}
Our reward model training pipeline is initialized with the Llama3.1-8B model~\citep{llama3} as its foundation. The process commences with supervised fine-tuning (SFT) on the UltraFeedback dataset to tailor the base model for our specific task requirements. Following the SFT phase, we proceed with reward model training, using the fine-tuned SFT model as the starting point. The comprehensive hyperparameter configurations for both the SFT and reward model training stages are systematically presented in \cref{table:rm_train_sft_rm_hyperparameters}.

\begin{tabular}{cc}
\label{exp_config}
    \begin{minipage}{.48\linewidth}
        \begin{table}[H]
        \captionsetup{font=small}
        \caption{
        \small The hyperparameter settings of SFT and reward model training.
        } 
        \label{table:rm_train_sft_rm_hyperparameters} 
        \centering 
        \begin{tabular}{@{}ll@{}}
        \toprule
        Hyperparameter              & Value   \\ \midrule
    Global Train Batch Size & 256 \\
    Micro Train Batch Size & 4 \\
    Train Epochs  & 1\\
    Max Total Length  & 1024\\
    DeepSpeed ZeRO stage & 3 \\
    Optimizer & Adam \\
    Learning Rate & 3e-6 \\
    Weight decay & 0.2 \\
    Scheduler Type & cosine\\
    Gradient clipping norm & 1.0 \\
    Dimension of $q$ & 256 \\
    Dimension of $k$ & 256 \\
    Offload & True \\
    \bottomrule
        \bottomrule
        \end{tabular}
        \end{table}
    \end{minipage} & 
    \begin{minipage}{.48\linewidth}
        \begin{table}[H]
        \captionsetup{font=small}
        \caption{
        \small The hyperparameter settings of Reinforce++ in RL training.} 
        \label{table:rm_train_rl_hyperparameters} 
        \centering 
        \begin{tabular}{@{}lll@{}}
        \toprule
        Hyperparameter              & Value   \\ \midrule
    Global Train Batch Size & 128 \\
    Micro Train Batch Size & 8 \\
    Global Rollout Batch Size & 1024 \\
    Micro Rollout Batch Size & 16 \\
    Train Epochs  & 1\\
    Max total length  & 1024\\
    Max prompt length & 512 \\
    Max generation length &512 \\
    DeepSpeed ZeRO stage & 2 \\
    Actor learning rate & 5e-7 \\
    Optimizer & Adam \\
    Gradient clipping norm & 1.0 \\
    KL coefficient & 0.01\\
    \bottomrule
         \bottomrule
        \end{tabular}
        \end{table}
    \end{minipage}
\end{tabular}

For experiments about reward model training, they are conducted on a server outfitted with AMD-EPYC9374F-32-Core Processor, 8 GeForce RTX 4090 GPUs, running Pop OS 22.04 LTS. For experiments about RL training, they are conducted on a server outfitted with 13th GenIntel(R)Core(TM)i9-13900K CPU, 4 NVIDIA A100 GPUs, runing Ubuntu 22.04.
\subsection{BoN}
\label{subsec:bon_details}
We implement Best-of-N (BoN) sampling to evaluate the performance of our reward models in inference-time alignment. Using the Llama-3.1-8B-Instruct model on the AlpacaEval benchmark \cite{alpaca_eval}, we generate response candidates with stochastic sampling parameters: temperature ($\tau=0.6$), top-k filtering ($\text{top-}\text{k}=50$), and nucleus sampling ($\text{top-}\text{p}=0.9$). For each prompt, we collect 32 independent responses. To analyze performance progression, we hierarchically select the best response by applying our trained reward model to subsets of $N \in \{2, 4, 8, 16, 32\}$ candidates, retaining the highest-scoring candidate for each $N$ configuration.

The selected responses are compared against baseline generations from the same model using greedy decoding ($\tau=0$, $\text{do}\_\text{sample}=\text{False}$). 

\subsection{Prompt for DeepSeek-V3 as Judge Model}
We adopt the DeepSeek-V3 as our automated judge model. The evaluation prompt is shown as below.

\newtcblisting{promptbox}{
    listing only,
    enhanced,
    breakable,
    colback=white,
    colframe=blue!50!black,
    sharp corners,
    boxrule=0.5pt,
    title={\textbf{Prompt used in AlpacaEval}},
    fonttitle=\bfseries,
    coltitle=black,
    before skip=10pt,
    after skip=10pt,
    attach boxed title to top left={xshift=5mm, yshift=-2mm},
    boxed title style={colback=blue!50!white, sharp corners},
    listing options={
        basicstyle=\ttfamily\small,
        breaklines=true,
        breakindent=0pt,
        showstringspaces=false,
        columns=fullflexible
    }
}

\begin{promptbox}
<|im_start|>system
You are a helpful assistant, that ranks models by the quality of their answers.
<|im_end|>

<|im_start|>user
I want you to create a leaderboard of different of large-language models. To do so,
I will give you the instructions (prompts) given to the models, and the responses of 
two models. Please rank the models based on which responses would be preferred by 
humans. All inputs and outputs should be python dictionaries.

Here is the prompt:
{
    "instruction": """{instruction}""",
}

Here are the outputs of the models:
[
    {
        "model": "model_1",
        "answer": """{output_1}"""
    },
    {
        "model": "model_2",
        "answer": """{output_2}"""
    }
]

Now please rank the models by the quality of their answers, so that the model with
rank 1 has the best output. Then return a list of the model names and ranks, i.e., 
produce the following output:
[
    {'model': <model-name>, 'rank': <model-rank>},
    {'model': <model-name>, 'rank': <model-rank>}
]

Your response must be a valid Python dictionary and should contain nothing else
because we will directly execute it in Python. Please provide the ranking that the
majority of humans would give.
<|im_end|>
\end{promptbox}

\subsection{RL Training}

In our reinforcement learning (RL) experiments, we maintain reproducibility by fixing the random seed to 1234 across all trials. The comprehensive hyperparameter configurations for RL training are systematically detailed in \cref{table:rm_train_rl_hyperparameters}.


We employ the AdamW optimizer with $\beta_1=0.9$ and $\beta_2=0.95$, augmented by three advanced memory optimization techniques: (1) ZeRO-2 stage optimization \cite{Zero-2} for distributed parameter management, (2) flash attention mechanisms \cite{Flashattn-2} for efficient attention computation, and (3) gradient checkpointing with CPU offloading to maximize GPU memory utilization.

The RL training protocol is configured with a fixed learning rate of $5 \times 10^{-7}$, micro batch sizes of 8 (training) and 16 (rollout), global batch sizes of 128 (training) and 1024 (rollout), and a uniform KL divergence penalty coefficient of 0.01 across all reward models.

For evaluation on the AlpacaEval benchmark \cite{alpaca_eval}, we use generation parameters including a temperature of 1.0, top-p sampling of 1.0, disabled top-k sampling, and a maximum of 512 new tokens. 
\section{More Experiment Results}
        

\begin{figure}[htbp]
    \centering
    \begin{minipage}{0.45\linewidth}
        \includegraphics[width=\linewidth, trim=0.0cm 0.2cm 0.5cm 1.0cm, clip]{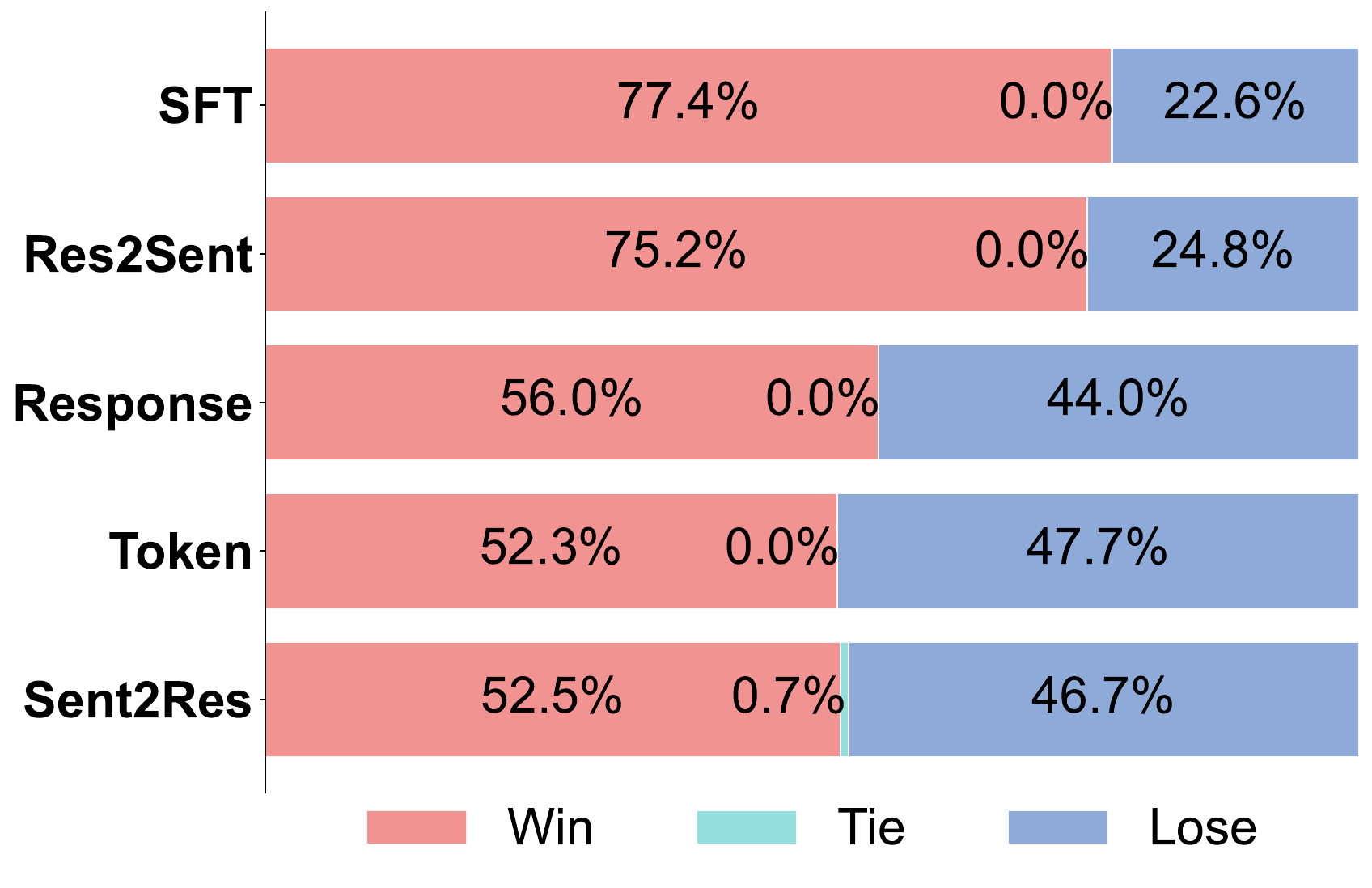}
        \caption{Win rate of our method against other baseline methods on AlpacaEval annotated by GPT-4o.}
        \label{fig:alpaca_eval_4o_a}
    \end{minipage}
    \hfill
    \begin{minipage}{0.45\linewidth}
        \includegraphics[width=\linewidth, trim=0.0cm 1.2cm 0.5cm 1.0cm, clip]{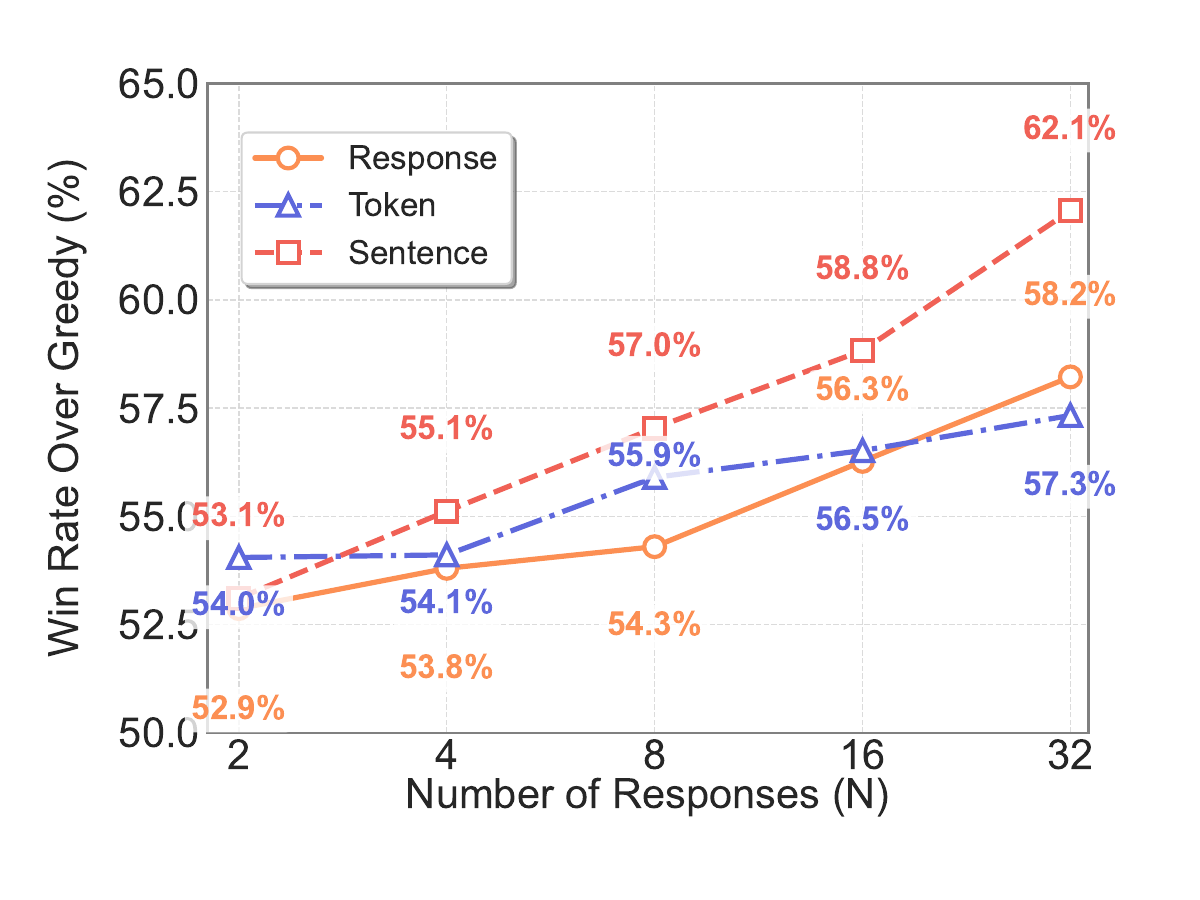}
        \caption{Win rate of BoN alignment with reward models over greedy decoding on AlpacaEval annotated by GPT-4o.}
        \label{fig:alpaca_eval_4o_b}
    \end{minipage}
\end{figure}
\vspace{10pt}

\begin{table}[htbp]
\centering
\small
\renewcommand{\arraystretch}{1.2} 
\setlength{\tabcolsep}{8pt} 
\label{tab:pairwise_accuracy}
\begin{tabular}{l|ccccc}
\toprule
\textbf{Method} & \textbf{Chat} & \textbf{Chat-Hard} & \textbf{Safety} & \textbf{Reasoning} & \textbf{Average} \\
\midrule
DA & 92.18 & 60.96 & 77.03 & 81.83 & 78.00 \\
VA    & 93.02 & 61.40 & 75.00 & 82.95 & 78.09 \\
VW & 94.69 & 59.43 & \textbf{81.49} & 83.58 & 79.80 \\
\textbf{ours} & \textbf{94.97} & \textbf{61.84} & 77.16 & \textbf{88.89} & \textbf{80.72} \\
\bottomrule
\end{tabular}
\caption{Pairwise Accuracy (\%) on RewardBench for Reward Models Trained with Different Aggregation Functions.}
\label{tab:agg_reward_bench}
\end{table}

\begin{table}[htbp]
\centering
\small
\begin{tabular}{l|ccc}
    \toprule
    \multirow{2}{*}{\textbf{Method}} & \multicolumn{3}{c}{\textbf{AlpacaEval}} \\
    \cmidrule(lr){2-4}
    & \textbf{LC (\%)} & \textbf{WR (\%)} & \textbf{Len.} \\
    \midrule
    DA       & 57.90 & 54.91 & 219 \\
    VA          & 59.53 & 70.75 & 388 \\
    VW       & 59.33 & 75.53 & 501 \\
    \textbf{Ours}       & \textbf{60.32} & \textbf{78.39} & 558 \\
    \bottomrule
\end{tabular}
    \caption{Comparison of aggregation methods on AlpacaEval with the Llama-3.2-3B-SFT model.}
    \label{tab:agg_alpaca}
\end{table}

\subsection{Using GPT-4o as the Judge Model}
In our Best-of-N (BoN) and reinforcement learning (RL) experiments, we evaluate model performance using DeepSeek-V3 as the primary judge model. This selection is based on its superior human agreement score (69.60\% vs. 69.17\% for GPT-4o), lower variance in judgments (12.50 vs. 14.62), and reduce tendency to favor longer responses (0.65 vs. 0.68). 
To further validate the robustness of our findings, we employ GPT-4o as an additional judge model, providing an independent assessment of the win rate. These complementary results are presented in \cref{fig:alpaca_eval_4o_a} and \cref{fig:alpaca_eval_4o_b}. We find that our method outperforms the baselines, which is consistent with the results obtained using DeepSeek-V3 as the judge model.
\subsection{The Impact of Aggregation Functions}
\label{form_agg}
The aggregation function employed in our method introduces an inductive bias into the training process, which has a significant impact on the overall performance. In \cref{agg_fn}, we investigate the impact of aggregation functions on performance. The specific forms of the aggregation functions are listed below.
\begin{itemize}
    \item \textbf{VW} : $r_\phi(x, y) = \sum_{i = 1}^{n_c}w_ir_\phi(x,c_1, \cdots, c_i)$.
    \item \textbf{VA} : $r_\phi(x, y) = \frac{1}{n_c}\sum_{i = 1}^{n_c}r_\phi(x, c_1, \cdots, c_i)$.
    \item \textbf{DA} : $r_\phi(x,y) = \frac{1}{n_c}\sum_{i = 1}^{n_c}\left(r_\phi(x, c_1, \cdots, c_i) - r_\phi(x, c_1, \cdots, c_{i - 1})\right)$.
\end{itemize}

\cref{tab:agg_reward_bench} reports the results on RewardBench and \cref{tab:agg_alpaca} reports the detailed win rates of different aggregation functions against SFT on the AlpacaEval for Llama-3.2-3B-SFT. It's observed our proposed aggregation functions have a better performance on RewardBench and AlpcalEval.
\begin{table}[htbp]
\centering
\small
\renewcommand{\arraystretch}{1.1}
\setlength{\tabcolsep}{5pt}
\makebox[\textwidth][c]{%
\begin{tabular}{l|ccc|cc}
\toprule
\textbf{SaT Model} & \multicolumn{3}{c|}{\textbf{AlpacaEval}} & \multicolumn{2}{c}{\textbf{Arena-Hard}} \\
& LC (\%) & WR (\%) & Len. & WR (\%) & Len. \\
\midrule
\texttt{sat-1l}       & \textbf{65.86} & 71.43 & 314 & 70.90 & 371 \\
\texttt{sat-9l}          & 61.79 & 77.64 & 554 & \textbf{77.00} & 525 \\
\texttt{sat-3l} & 58.77 & \textbf{78.39} & 623 & 76.20 & 638 \\
\bottomrule
\end{tabular}
}
\caption{The win rates(WR) and Length-Control win rates(LC) of models trained with different segment models on AlpacaEval2 and Arena-Hard benchmarks.}
\label{tab:agg_alpaca_arena}
\end{table}

\subsection{The Impact of Segment Models}
In \cref{tab:agg_alpaca_arena}, we show the results using three different SaT models for segmentation. While \texttt{sat-9l} exhibits the best overall performance, it significantly increases the RL training time. \texttt{sat-3l}, in contrast, provides a better trade-off between performance and training time.
\section{Examples of Generated Response}
\label{appendix:sample_of_response}

In this section, we study cases selected from AlpacaEval, from which we observe distinct behavioral patterns across reward modeling approaches. The sentence-level reward model yields responses that strike a balance between depth and conciseness: compared to response-level optimization (which produces overly generic outputs), it incorporates richer contextual details, while avoiding the verbosity and factual inconsistencies prevalent in token-level optimization. In \cref{fig:sample_alpaca_eval_1} and \cref{fig:sample_alpaca_eval_2}, we present specific examples that illustrate these differences, highlighting the superiority of the sentence-level reward model in enhancing the final policy's training performance.

\begin{figure*}[!p]
    \label{fig:case1}
    \small 
    \centering
    \begin{spacing}{0.9} 
    \begin{tabular}{@{}p{0.98\linewidth}@{}}
    \toprule \toprule
    \textcolor[HTML]{3078BE}{\textbf{Instruction: Are there any weird Christmas traditions?}}  \\ \toprule \toprule  
    
    \textbf{Response:}  There are indeed many unique Christmas traditions in various regions across the world. Some might be more well-known, like the German tradition of leaving out nuts and hay for mice, \textcolor{red}{(incorrect, not a known Christmas tradition)} while others might be less common, like the Scottish custom of singing carols while carrying a broom! While these customs are interesting, it's important to remember that all traditions should foster a spirit of warmth, unity, and goodwill.
    \\
    \hline 

    \textbf{Token:}  Yes, many cultures have their unique traditions associated with Christmas, some of these traditions may sound somewhat odd to us, such as the Christmas stocking exchange originated in England, which involves adults filling up their own stockings at the Christmas celebration. \textcolor{red}{(incorrect, stockings are typically filled for children, not adults)} While these traditions may not be everyone's personal favorites, they certainly add a layer of cultural diversity and charm to the festive season. Remember, traditions aren't about the materialism of gifts or decorations, but about passing down values, emotions, and memories that hold sentimental significance to the people involved. Some traditions, while not quite as well-known, are quite interesting, including the tradition of leaving cookies for jolly old Saint Nicholas in the hope of receiving gifts in return, or putting up the Christmas tree after Saint Nicholas Day on December 6th! These traditions showcase the rich human experiences and perspectives around Christmas celebrations worldwide. So while some traditions might appear a bit unusual to us, they're also valuable in contributing to the diverse tapestry of holiday festivities around the world. Next time around, why not explore more about these unexpected yet meaningful holiday customs, they might surprise and delight you. For more insight, a recommended book to explore is "Christmas Around the World" by Martha Day Zschock, which showcases the multifaceted traditions of various regions and cultures. 
    \\
    \hline  
    
    \textbf{Sentence:} Yes, many cultures have their unique traditions associated with Christmas, some of these traditions may sound somewhat odd to us, such as the Christmas stocking exchange originated in England, which involves adults filling up their own stockings at the Christmas celebration. There are also numerous traditions associated with Christmas tree decorating which can vary from place to place. For instance, Germany has a distinct tradition of hanging wreaths adorned with candles called Lichterketten on their Christmas trees, \textcolor[HTML]{1D9942}{(accurate and specific example)} while in Scandinavia, they typically wear masks during their Christmas parties. In Ireland, it's common to serve mulled wine, \textcolor[HTML]{1D9942}{(culturally accurate detail)} while in Portugal, the use of a nativity scene called a presépe is widespread. Different versions of The Twelve Days of Christmas are also very famous traditions in some places. These are just some examples of unique and distinctive Christmas traditions across geographical locations around the world. Remember, it's all about celebrating the spirit of Christmas with love and unity. How about starting an engaging conversation about these traditions with a family member, a friend, or a colleague today? \textcolor[HTML]{1D9942}{(engaging and informative conclusion)}
    \\
    \bottomrule
\end{tabular}
\end{spacing}
\caption{Comparison of responses to a question about unusual Christmas traditions. In the examples, correct and accurate information is explained in dark green, while incorrect or misleading details are marked in red. \textbf{Sentence} provides a detailed, accurate, and engaging overview of unique traditions across cultures, such as Germany's Lichterketten and Ireland's mulled wine, demonstrating its superiority. In contrast, \textbf{Response} incorrectly describes a German tradition of leaving nuts and hay for mice, and \textbf{Token} inaccurately states that adults fill their own stockings in England. \textbf{Sentence}'s response is both informative and culturally precise, making it the most reliable and comprehensive answer.}
\label{fig:sample_alpaca_eval_1}
\end{figure*}

\begin{figure*}[htbp]
    \label{case2}
    \small 
    \centering
    \begin{spacing}{0.9} 
    \begin{tabular}{@{}p{0.98\linewidth}@{}}
    \toprule \toprule
    \textcolor[HTML]{3078BE}{\textbf{Instruction: Write a C++ function that takes a reference to a std::string containing markdown formatted text and returns a std::string containing html formatted text.}}  \\ \toprule \toprule  
    
    \textbf{Response:}  Here's some sample code that accomplishes what you're asking for:

\begin{verbatim}
std::string escapeMarkdown(const std::string& markdown_code) {
  return std::string(markdown_code.begin(), markdown_code.end());
}
\end{verbatim}
\textcolor{red}{(Fails to convert Markdown to HTML)}
\begin{verbatim}
int main() {
  std::string markdown_body = "# Welcome To My Website\n\nThis website was "
                              "created as a part of a course. It mainly "
                              "features content related to my course projects "
                              "and technical skills acquired during the course. "
                              "Let's dive in!";
  std::string html_rendered_body = escapeMarkdown(markdown_body);
  return 0;
}
\end{verbatim}
    \\
    \hline 

    \textbf{Token:}  Here's a possible implementation of a function that takes a markdown formatted string as input and returns its html counterpart:
    \begin{verbatim}
    #include <iostream>
    #include <string>
    #include <cstring>
    #include <cctype>
    std::string markdown2html(const std::string& md) {
        std::ostringstream oss;

        // Replace asterisks with actual bold text
        std::string re = "[b][u]([0-9a-z]*)[/u][/b]";
        std::string final_re = ".*?";
        oss << "<b><u>";
        while (md.substr(0).find_first_of(" *", final_re, re)!= std::string::npos)
        {
            int idx = md.find_first_of(" *", final_re, re);
            const std::string chunk = md.substr(0, idx);
            const std::string newChunk = chunk.replace(chunk.begin(), chunk.end(), "<b><u>");
            md = md.substr(idx + 1); // Remove the original character(s) from the md string
            oss << newChunk << " ";
        }
        // Replace underscores with actual italicized text
        oss << "<sub>[i]";
        re = "[i]";
        final_re = ".*?";
        oss << "<u>*([0-9a-z]*)*</u>";
        oss << "</sub>";
        while (md.substr(0).find_first_of(" *", final_re, re)!= std::string::npos)
        {
            int idx = md.find_first_of(" *", final_re, re);
            const std::string chunk = md.substr(0, idx);
            const std::string newChunk = chunk.replace(chunk.begin(), chunk.end(), "<b><u>");
            md = md.substr(idx + 1); // Remove the original character(s) from the md string
            oss << newChunk << " ";
        }
        return oss.str();
    }
    \end{verbatim}
    \textcolor{red}{(Incorrect implementation and overly complex)}
    \\
    \hline  
    
    \textbf{Sentence:} Sure, here is an example of how we can create a generic function that converts Markdown to HTML:

\begin{verbatim}
std::string convert_md_to_html(const std::string& input_text) {
   std::stringstream html_stream;
   html_stream << "<p>" << input_text << "</p>";
   return html_stream.str();
}
\end{verbatim}
 \textcolor[HTML]{1D9942}{(A concise and simple implementation)}
    \\
    \bottomrule
\end{tabular}
    \end{spacing}
\caption{Comparison of responses to a C++ Markdown-to-HTML conversion task. \textbf{Sentence} provides a concise and functional solution, correctly wrapping the input text in HTML paragraph tags, demonstrating clarity and simplicity. In contrast, \textbf{Response} fails to convert Markdown to HTML, merely escaping the input, while \textbf{Token} presents an overly complex and incorrect implementation, with excessive and non-functional code.  }
\label{fig:sample_alpaca_eval_2}
\end{figure*}

\clearpage

\end{document}